\documentclass[10pt,twocolumn,letterpaper]{article}

\usepackage{iccv}
\usepackage{times}
\usepackage{epsfig}
\usepackage{graphicx}
\usepackage{amsmath}
\usepackage{amssymb}
\usepackage{booktabs}
\usepackage{bbold}
\usepackage{xcolor}
\usepackage{caption}
\usepackage{multirow}
\usepackage{soul}
\usepackage{bbm}
\usepackage{wrapfig}
\usepackage{xcolor}
\usepackage{enumitem}
\usepackage[title]{appendix}

\usepackage[pagebackref=true,breaklinks=true,letterpaper=true,colorlinks,bookmarks=false]{hyperref}

\iccvfinalcopy

\ificcvfinal\fi

\begin{document}


\title{A-STAR: Test-time \underline{A}ttention \underline{S}egrega\underline{t}ion \underline{a}nd \underline{R}etention \\ for Text-to-image Synthesis}

\author{Aishwarya Agarwal, Srikrishna Karanam, K J Joseph, Apoorv Saxena,\\ Koustava Goswami, and Balaji Vasan Srinivasan\\
Adobe Research, Bengaluru India \\
{\tt \scalebox{.7}{\{aishagar,skaranam,josephkj,apoorvs,koustavag,balsrini\}@adobe.com}}
}

\twocolumn[{
\renewcommand\twocolumn[1][]{#1}%
\maketitle
\begin{center}
 \centering
 \captionsetup{type=figure}
 \includegraphics[width=1.0\textwidth]{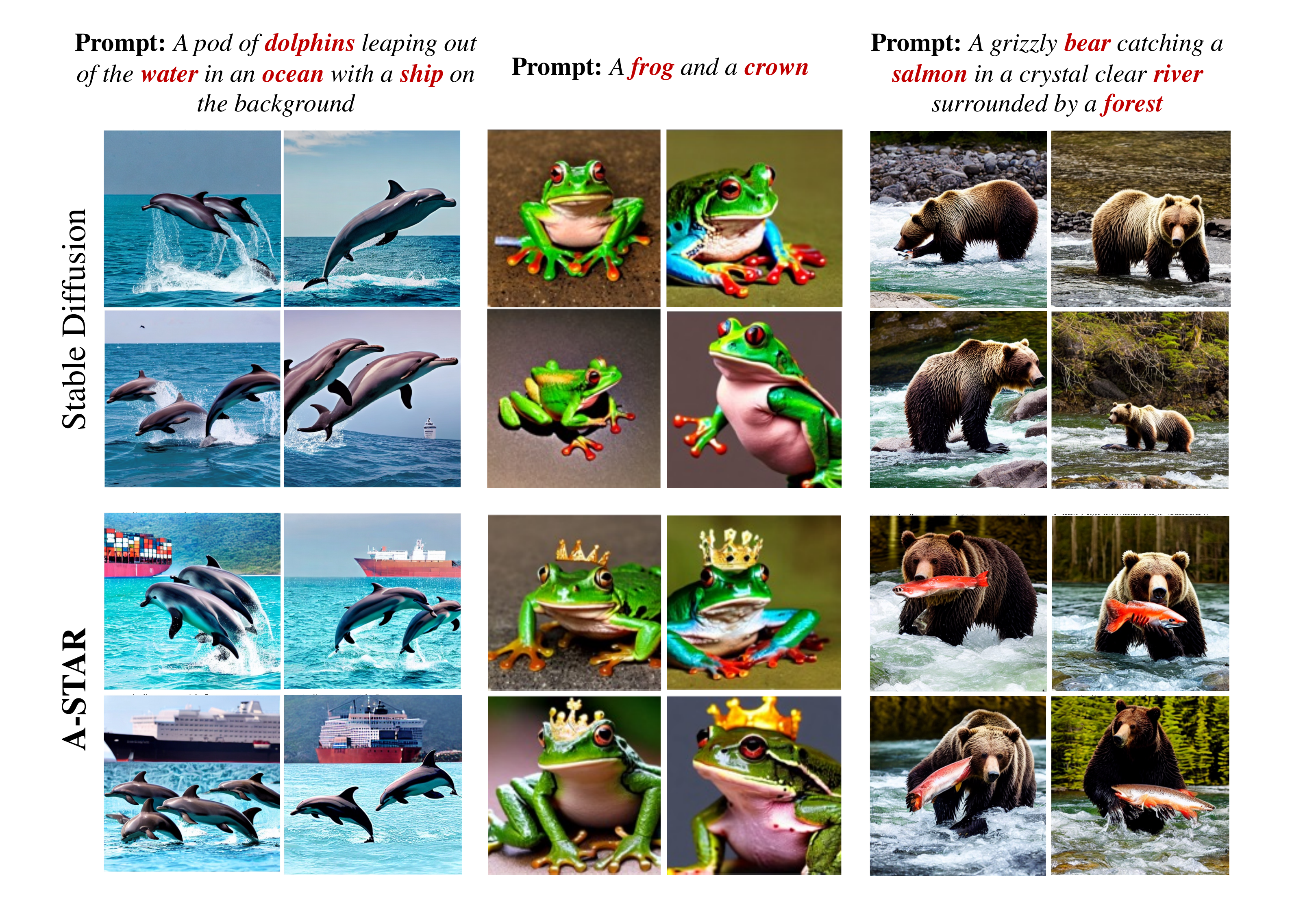}
 \vspace{-34pt}

 \caption{Despite their remarkable ability to generate plausible images from text descriptions, diffusion models fail to be faithful to multiple concepts in the input text. We identify the issues causing this pitfall, and propose a training-free method to fix them. We propose two new loss functions, attention segregation loss and attention retention loss, that only require test time optimization to drive the diffusion process and produce substantially improved generation results. We can note from these results that our method captures all key concepts in the input prompt as opposed to baseline Stable Diffusion \cite{rombach2022high}.}
 \label{fig:teaser_qual}
\end{center}
}]

\maketitle
\ificcvfinal\thispagestyle{empty}\fi

\begin{abstract}

While recent developments in text-to-image generative models have led to a suite of high-performing methods capable of producing creative imagery from free-form text, there are several limitations. By analyzing the cross-attention representations of these models, we notice two key issues. First, for text prompts that contain multiple concepts, there is a significant amount of pixel-space overlap (i.e., same spatial regions) among pairs of different concepts. This eventually leads to the model being unable to distinguish between the two concepts and one of them being ignored in the final generation. Next, while these models attempt to capture all such concepts during the beginning of denoising (e.g., first few steps) as evidenced by cross-attention maps, this knowledge is not retained by the end of denoising (e.g., last few steps). Such loss of knowledge eventually leads to inaccurate generation outputs. 

To address these issues, our key innovations include two test-time attention-based loss functions that substantially improve the performance of pretrained baseline text-to-image diffusion models. First, our \textbf{attention segregation loss} reduces the cross-attention overlap between attention maps of different concepts in the text prompt, thereby reducing the confusion/conflict among various concepts and the eventual capture of all concepts in the generated output. Next, our \textbf{attention retention loss} explicitly forces text-to-image diffusion models to retain cross-attention information for all concepts across all denoising time steps, thereby leading to reduced information loss and the preservation of all concepts in the generated output. We conduct extensive experiments with the proposed loss functions on a variety of text prompts and demonstrate they lead to generated images that are significantly semantically closer to the input text when compared to baseline text-to-image diffusion models. 
\end{abstract}

\section{Introduction}
\label{sec:intro}

The last few years has seen a dramatic rise in the capabilities of text-to-image generative models to produce creative image outputs conditioned on free-form text inputs. While the recent class of pixel \cite{ramesh2022hierarchical,saharia2022photorealistic} and latent \cite{rombach2022high} diffusion models have shown unprecedented image generation results, they have some key limitations. First, as noted in prior work \cite{feng2022training,wang2022diffusiondb,chefer2023attend}, these models do not always produce a semantically accurate image output, consistent with the text prompt. As a consequence, there are numerous cases where not all subjects of the input text prompt are reflected in the model's generated output. For instance, see Figure~\ref{fig:teaser_qual} where Stable Diffusion \cite{rombach2022high} omits \textit{ship} in the first column, \textit{crown} in the second column, and \textit{salmon} in the third column.

\begin{figure}
    \centering
    \includegraphics[width = 1.01\linewidth]{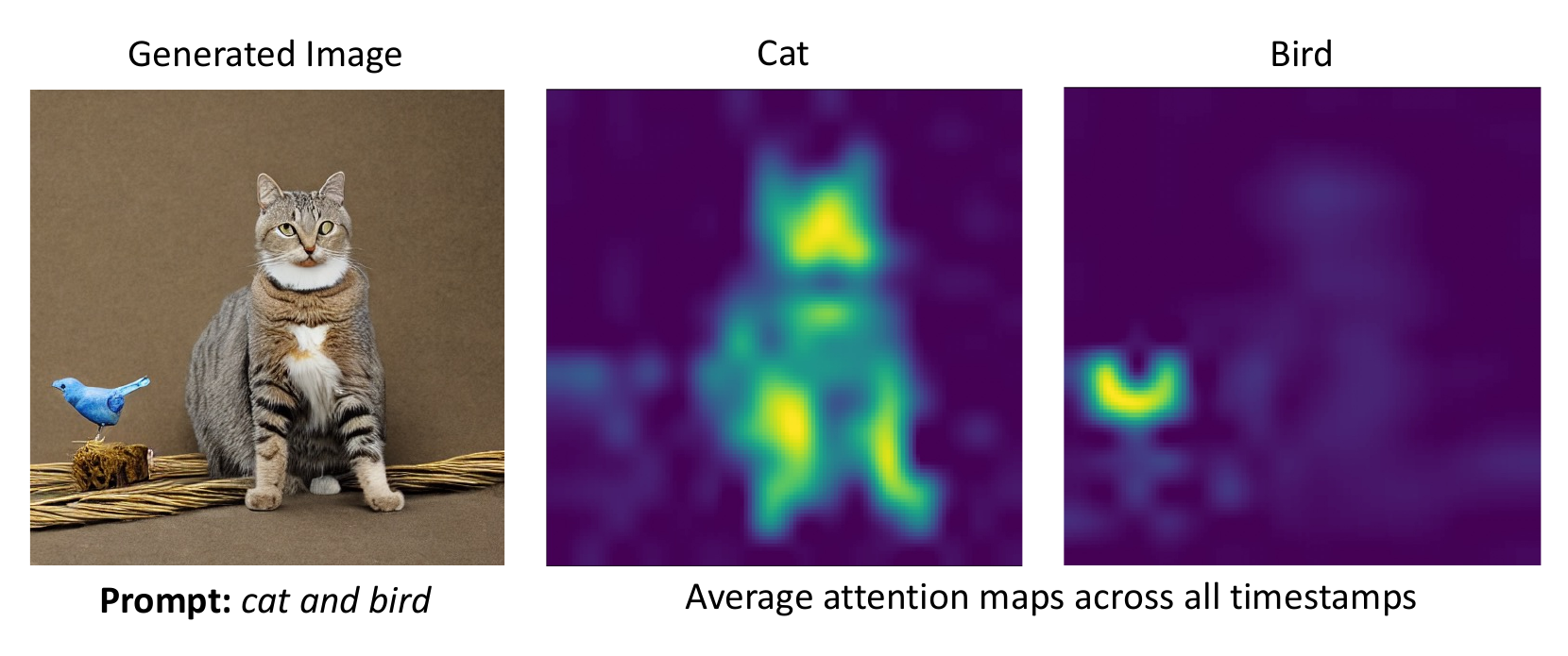}
    \caption{Cross-attention maps for \textit{cat} and \textit{bird}.}
    \label{fig:cross_attention}
\end{figure}

To understand the reasons for these issues, we compute and analyze the cross-attention maps produced by these models during each denoising time step. Specifically, as noted in prior work \cite{hertz2022prompt}, the interaction between the input text and the generated pixels can be captured in attention maps that explicitly use both text features and the spatial image features at the current time step. For instance, see Figure~\ref{fig:cross_attention} that shows per-subject-token cross-attention maps, where one can note high activations eventually lead to expected outputs. By analyzing these maps, we posit we can both understand why models such as Stable Diffusion fail (as in Figure~\ref{fig:teaser_qual}) as well as propose ways to address the issues.

\begin{figure}
    \centering
    \includegraphics[width = 1.01\linewidth]{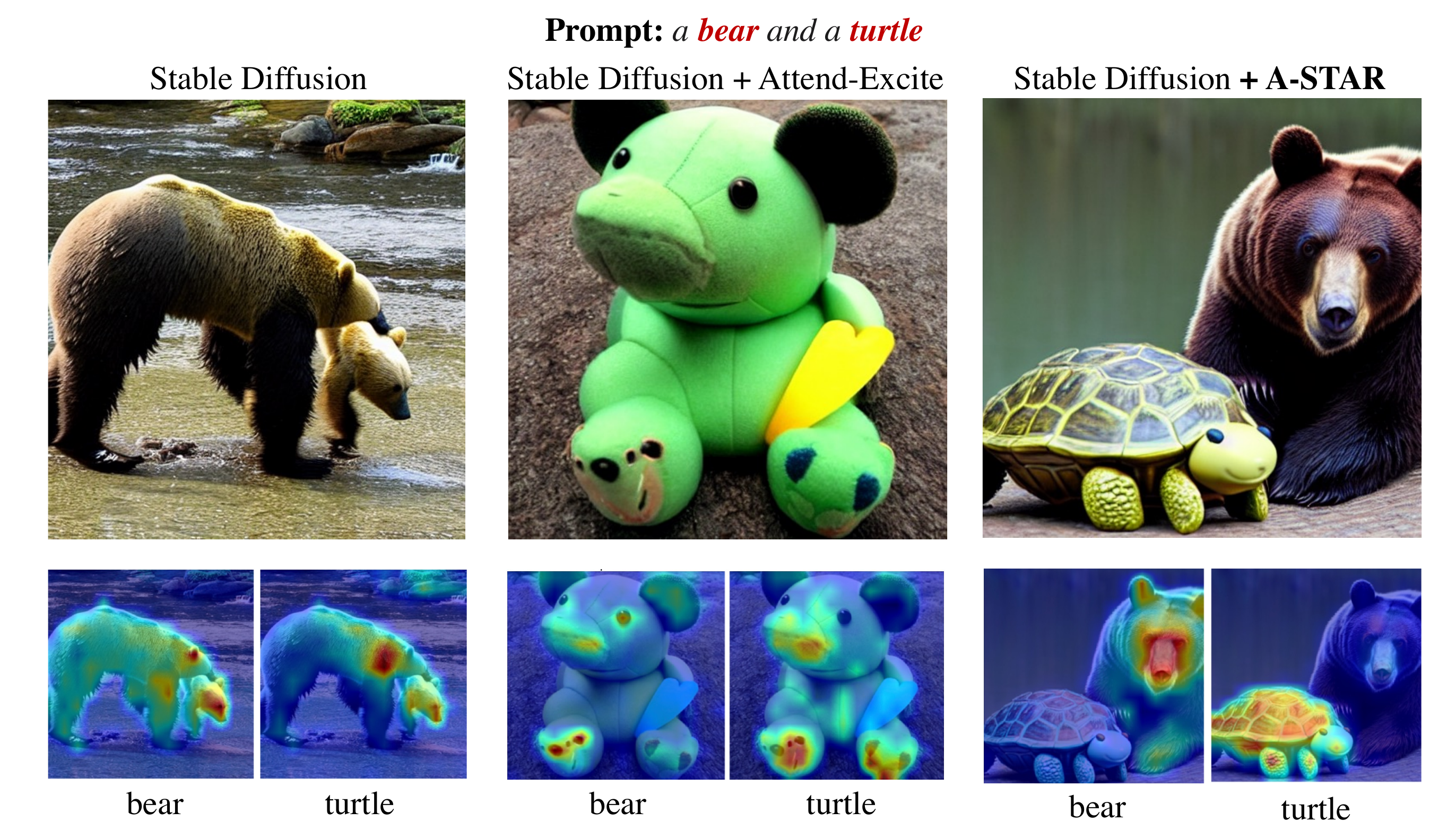}
    \vspace{-20pt}
    \caption{Our proposed method reduces the overlap between various concepts' attention maps, leading to reduced confusion/conflicts and improved generation results when compared to baseline models. In this example, one can note much less overlap in the high-response regions for \textit{bear} and \textit{turtle} with our method when compared to baselines.}
    \label{fig:attention_overlap}
\end{figure}

Based on our observations of these cross-attention maps, we notice two key issues with existing models such as Stable Diffusion \cite{rombach2022high} that lead to incorrect generation outputs. First, in cases that involve multiple subjects in the text prompt, we notice the presence of a significant amount of \textit{overlap} between each subject's cross-attention map. Let us consider the example in Figure~\ref{fig:attention_overlap}. We compute the average (across all denoising steps) 
cross-attention maps for \textit{bear} and \textit{turtle} and notice there is significant overlap in the regions that correspond to high activations. We conjecture that because both \textit{bear} and \textit{turtle} are highly activated in \textit{the same pixel regions}, the final generated image is unable to distinguish between the two subjects and is able to pick only one of the two. Note that even exciting the regions as done in Attend-Excite \cite{chefer2023attend} does not help to alleviate this issue. We call this issue with existing models as \textbf{attention overlap}.

\begin{figure}
    \centering
    \includegraphics[width = \linewidth]{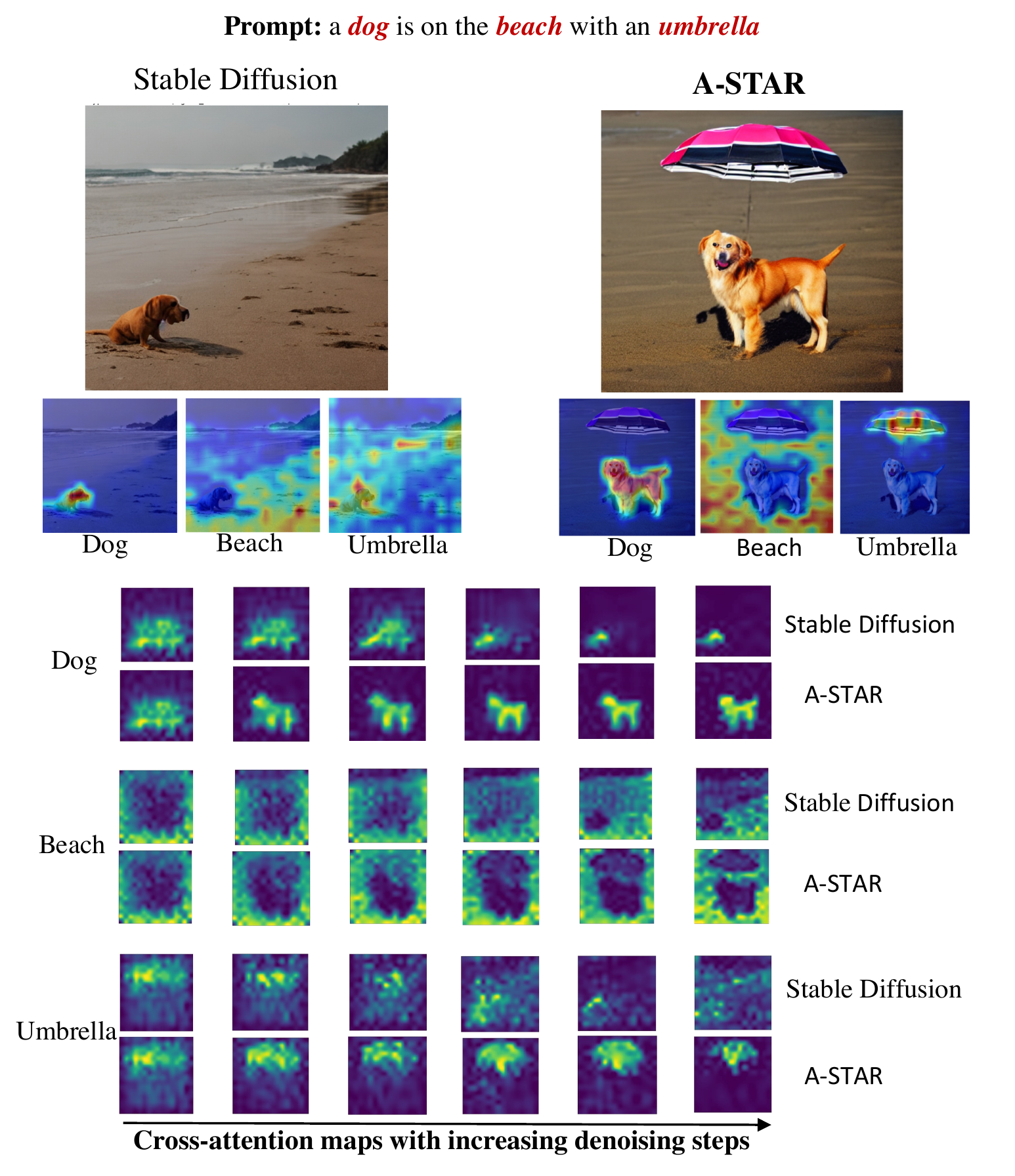}
    \caption{Our proposed method ensures information retention for all concepts across all denoising steps, leading to improved generation. In this example, one can note pixel activations for \textit{dog}, \textit{beach}, and \textit{umbrella} are retained across all steps with our method when compared to the baseline.}
    \label{fig:attention_retention}
\end{figure}

Our next observation is related to an issue we call \textbf{attention decay} of text-to-image diffusion models. Let us consider the result in Figure~\ref{fig:attention_retention} where we show the cross-attention maps for \textit{dog}, \textit{beach}, and \textit{umbrella} across multiple denoising time steps for the Stable Diffusion model \cite{rombach2022high}. One can note that in the beginning of the diffusion process (e.g., steps 1-3), cross-attention maps for all three concepts were highly activated but this was \textit{not retained} by the time we reach the end of all the diffusion steps. In the end (see last column in the figure), one can note the pixel regions that were initially highly activated for these concepts are now either very sparsely activated or not activated at all. This suggests while the Stable Diffusion model is trying to capture all concepts in the input prompt during the early stages of diffusion, it is not able to retain this knowledge. This non-retention of information by the end of the diffusion process leads to the model missing out various parts of the text input in the generated output. For instance, in Figure~\ref{fig:attention_retention}, one can note \textit{umbrella} is very sparsely activated at the end (even though this was not the case in step 1), leading the Stable Diffusion's \cite{rombach2022high} generated output missing it. Please note that both these issues of attention overlap  and decay are prevalent in many cases and we illustrate the same in supplementary material due to space constraints. 

To address the aforementioned issues, we propose two new loss functions that only require inference-time optimization and no retraining of the base text-to-image diffusion models. First, our \textbf{attention segregation loss} tackles the attention overlap issue noted above by explicitly minimizing the overlap of high-response regions in the cross-attention maps of all concept pairs. Our key insight here is by explicitly \textit{segregating} the pixel regions that are highly activated for a pair of concepts, we ensure the model captures knowledge about both concepts, thereby generating both in the final output at the end of the denoising process. Next, our \textbf{attention retention loss} tackles the attention decay issue above by explicitly ensuring information retention across denoising time steps. We realize this by computing a mask for each concept's cross-attention map from the previous time step and ensuring the highly activated regions in the current time step's attention map is consistent with this mask. Our key insight here is by retaining highly-activated pixel regions for each subject across the entire denoising process, we explicitly equip the text-to-image model with the ability to retain all relevant knowledge by the end of denoising, thereby leading to improved generations. We show some results with our proposed losses in Figures~\ref{fig:attention_overlap} and~\ref{fig:attention_retention}. In Figure~\ref{fig:attention_overlap}, our method reduces the overlap between the \textit{bear} and \textit{turtle} attention maps, leading to improved generation when compared to baseline Stable Diffusion. Note that our method gives better results when compared to Attend-Excite \cite{chefer2023attend} since this technique only ensures all attention maps are activated but does not explicitly account for any overlap issues that we have identified. Next, in Figure~\ref{fig:attention_retention}, as can be seen from the progression of the attention maps, our method is able to retain high-response regions across the denoising steps, resulting in the final generation capturing all three concepts as opposed to baseline Stable Diffusion. 

\vspace{10pt}
\noindent To summarize, our key contributions are below:
\begin{itemize}
[leftmargin=*, noitemsep]
    \item We identify two key issues with existing text-to-image diffusion models, attention overlap and attention decay, that lead to semantically inaccurate generations like in Figure~\ref{fig:teaser_qual}.
    \item We propose two new loss functions called attention segregation loss and attention retention loss to explicitly address the above issues. These losses can be directly used during the test-time denoising process without requiring any model retraining. 
    \item The attention segregation loss minimizes the overlap between every concept pair's cross-attention maps whereas the attention retention loss ensures information for each concept is retained across all the denoising steps, leading to substantially improved generations when compared to baseline diffusion models without these losses. We conduct extensive qualitative and quantitative evaluations to establish our method's impact when compared to several baseline models.
    
\end{itemize}

\section{Related Work}
Before the emergence of large-scale diffusion models for conditional image synthesis, much effort was expended in using generative adversarial networks \cite{isola2017image,zhu2017unpaired,zhu2017toward,park2019semantic,karras2019style} and variational autoencoders \cite{huang2018introvae} for either conditional or unconditional image synthesis. With text-conditioned image synthesis having many practical applications, there was also much recent work in adapting generative adversarial networks for this task \cite{tao2022df, xu2018attngan, ye2021improving, zhang2021cross, zhu2019dm}. However, with the dramatic recent success of diffusion models \cite{nichol2021glide, rombach2022high, saharia2022photorealistic, ramesh2022hierarchical}, there have been a large number of efforts in very quick time to improve them. In addition to methods such as classifier-free guidance \cite{ho2022classifier}, there were also numerous efforts in the broad area of prompt engineering \cite{liu2022design, wang2022diffusiondb, witteveen2022investigating, hao2022optimizing} to adapt prompts so that the generated image satisfied certain desired properties. Other recent efforts also seek to customize these diffusion models \cite{yang2022reco,li2023gligen,zeng2022scenecomposer,brooks2022instructpix2pix, mokady2022null} to user inputs.

However, as discussed in Section~\ref{sec:intro}, existing text-to-image diffusion models are not able to capture all concepts in the input prompt, leading to semantically undesirable outputs. There have been some recent efforts to address this issue. In Liu et al. \cite{liu2022compositional}, the authors proposed a composition of diffusion models to generate the final output. However, this method often fails to generate realistic compositions and is limited to specific object properties. In Chefer et al. \cite{chefer2023attend}, the authors proposed to manipulate cross-attention maps by maximizing the activations of the most neglected concepts in the generated outputs. While this does take a step towards addressing the issue above, it fails in several cases (see Figure~\ref{fig:attention_overlap} and~\ref{fig:mainQualComparisonSOTA}) because attention maximization does not necessarily ensure all concepts are captured. As we discussed in Section~\ref{sec:intro}, overlap between two concepts' cross-attention maps and the non-retention of high activations over time are two critical issues that impact the final generations. We address these issues with new test-time loss functions that explicitly reduce attention overlap and ensure high-activation knowledge retention across denoising steps, leading to improvements over not only base models like Stable Diffusion \cite{rombach2022high} but also add-ons like Chefer et al. \cite{chefer2023attend}. 

\section{Approach}

\subsection{Latent Diffusion Models and Cross Attention}

\begin{figure}
    \centering
    \includegraphics[scale=0.41]{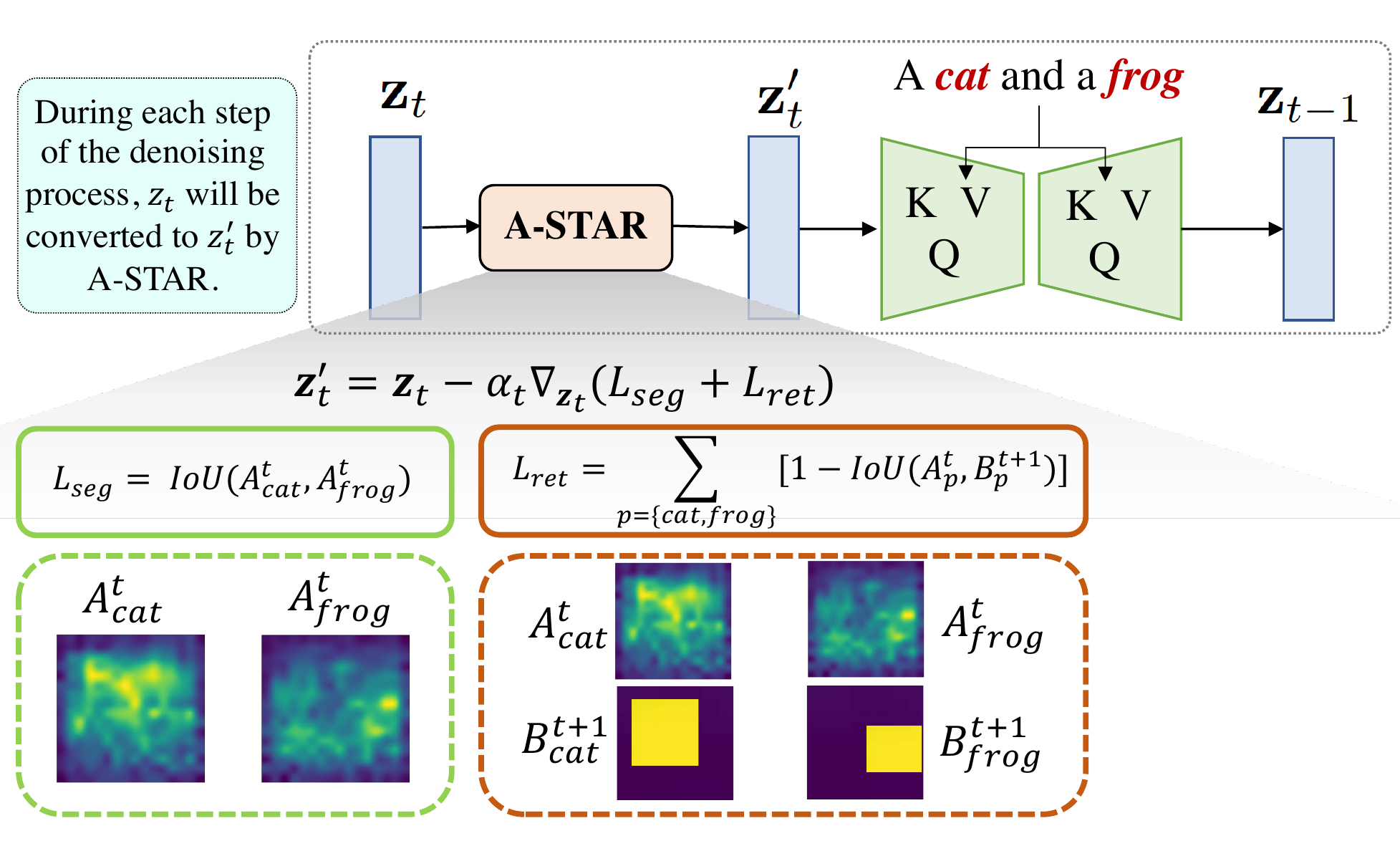}
    \caption{A visual illustration of the proposed A-STAR algorithm. See Eq.~\ref{eq:attn_segregation} for $L_\text{seg}$ and Eq.~\ref{eq:attn_retention} for $L_\text{ret}$.}
    \vspace{-12pt}
    \label{fig:placeholder}
\end{figure}

We start with a brief review of latent diffusion models (LDMs) and the associated cross-attention computation mechanism. LDMs comprise an encoder-decoder pair and a separately trained denoising diffusion probabilistic model (DDPM). In Rombach et al. \cite{rombach2022high}, the encoder-decoder pair is a standard variational autoencoder where an image $\mathbf{I} \in \mathcal{R}^{W\times H \times 3}$ is encoded to a latent code $\mathbf{z} = \mathbf{E}(\mathbf{I}) \in \mathcal{R}^{h\times w \times c}$ of much smaller spatial resolution (when compared to $\mathbf{I}$) using the encoder $\mathbf{E}$. A decoder $\mathbf{D}$ is trained to reconstruct the image $I \approx \mathbf{D}(\mathbf{z})$. The DDPM operates on the learned latent representations of the autoencoder (regularized using standard KL-type losses \cite{kingma2013auto,van2017neural}) in a series of denoising steps. In each step $t$, given the current latent code $\mathbf{z}_{t}$, the DDPM is trained to produce a denoised verison $\mathbf{z}_{t-1}$. This process can be conditioned using external conditioning factors, and this typically is the output of a text encoder $\mathbf{L}$ (e.g., CLIP \cite{radford2021learning}, T5 \cite{raffel2020exploring}). Given the input text prompt $p$'s encoding $\mathbf{L}(p)$ using the text encoder $\mathbf{L}$, the DDPM $\epsilon_{\mathbf{\Theta}}$, parametrized by $\mathbf{\Theta}$, is trained to optimize the following loss:
\begin{equation}
\mathbb{E}_{\mathbf{z}\sim \mathbf{E}(\mathbf{I}),p,\epsilon\sim\mathcal{N}(0,1),t}[\|\epsilon - \epsilon_{\mathbf{\Theta}}(\mathbf{z}_{t}, \mathbf{L}(p), t)\|] 
\end{equation}
Once the models (both autoencoder and DDPM) are trained, generating an image involves getting the text encoding of the input prompt $\mathbf{L}(p)$, sampling a latent code $\mathbf{z}_{T}\sim \mathcal{N}(0,1)$, running $T$ denoising steps using $\epsilon_{\mathbf{\Theta}}$ to obtain $\mathbf{z}_{0}$, and finally decoding using $\mathbf{D}$ to get $\mathbf{I}^{'}=\mathbf{D}(\mathbf{z}_{0})$.

In practice \cite{rombach2022high}, the $\epsilon_{\mathbf{\Theta}}$ model is implemented using the UNet architecture \cite{ronneberger2015u} with both self- and cross-attention layers. The cross-attention layers is where explicit text infusion happens using cross-attention \cite{vaswani2017attention} between projections of both $\mathbf{L}(p)$ and $\mathbf{z}_{t}$. As shown in prior work \cite{chefer2023attend,hertz2022prompt}, this results in a set of cross-attention maps $\mathbf{A}_t \in \mathbb{R}^{r \times r \times N}$ ($r=16$ from Hertz et al. \cite{hertz2022prompt}) at each denoising step $t$ for each of $N$ tokens (tokenized using $\mathbf{L}$'s tokenizer) in the input prompt $p$. From Figure~\ref{fig:cross_attention}, the cross attention map of cat and bird is indeed attending to the corresponding spatial location.

\subsection{A-STAR: \underline{A}ttention \underline{S}egrega\underline{t}ion \underline{a}nd \underline{R}etention}
\label{sec:asr}

As noted in Section~\ref{sec:intro}, we identify two key issues, attention overlap and attention decay, with existing models that result in semantically incorrect generations. Here, we first discuss our intuition that leads to these issues and then describe our proposed solutions to alleviate them. Let us consider the prompt \textit{a cat and a dog}. Using this input, we generate images using base Stable Diffusion \cite{rombach2022high} with varying seeds. In more than $80\%$ of the seeds, we notice either cat or dog were missing (see Fig~\ref{fig:baseSD_intuition}). 

For a very small number of seeds, we notice both cat and dog showing up in the final result (see bottom-right in Fig~\ref{fig:baseSD_intuition}). This suggests that while the DDPM model has all the semantic information it needs, \textit{the path it takes} to the final denoising result affects the generated image (and based on this analysis and results in Figure~\ref{fig:teaser_qual}, in most cases it ends up taking an undesirable path). To understand why, we look at the cross-attention maps that led to results in Figures~\ref{fig:attention_overlap} and~\ref{fig:attention_retention}.

In Figure~\ref{fig:attention_overlap}, we notice at a particular step in the denoising process, there is significant \textit{overlap} in the highly-activated pixel regions for both \textit{bear} and \textit{turtle} (i.e., high activations in same local image regions). Put another way, this is like saying the DDPM is considering putting both \textit{bear} and \textit{turtle} in the same local regions in the final generated image, leading to a clear case of confusion. As can be seen from the result (column 1), this is indeed the case, with only bear (no turtle) in the final generation. We call this issue attention overlap. In Figure~\ref{fig:attention_retention}, we notice that the DDPM has information on all concepts during the beginning of the denoising process, it is unable to retain this knowledge as denoising proceeds. See the cross-attention maps from Stable Diffusion in column 1, where one can note \textit{dog}, \textit{beach}, and \textit{umbrella} all have high activations at step T but this is lost by the time DDPM reaches $t=0$ (only \textit{dog} and \textit{beach} have high activations whereas \textit{umbrella} is lost). Consequently, the image decoded with $\mathbf{z}_{0}$ does not have all the concepts (see base Stable Diffusion result where only dog and beach show up). We call this issue attention decay. Our intuition and the results in Figures~\ref{fig:attention_overlap} and~\ref{fig:attention_retention} suggests if we are able to correct the issues above with these intermediate representations, we will be able to guide the DDPM denoising process in the right direction that eventually gives a $\mathbf{z}_{0}$ that can be decoded into a semantically-expected generation. This is where our contributions lie with two new test-time (no model retraining) loss functions we discuss next. We visually summarize our proposed method in Figure~\ref{fig:placeholder}.

\begin{figure*}
    \centering
    \includegraphics[width = 0.98\linewidth]{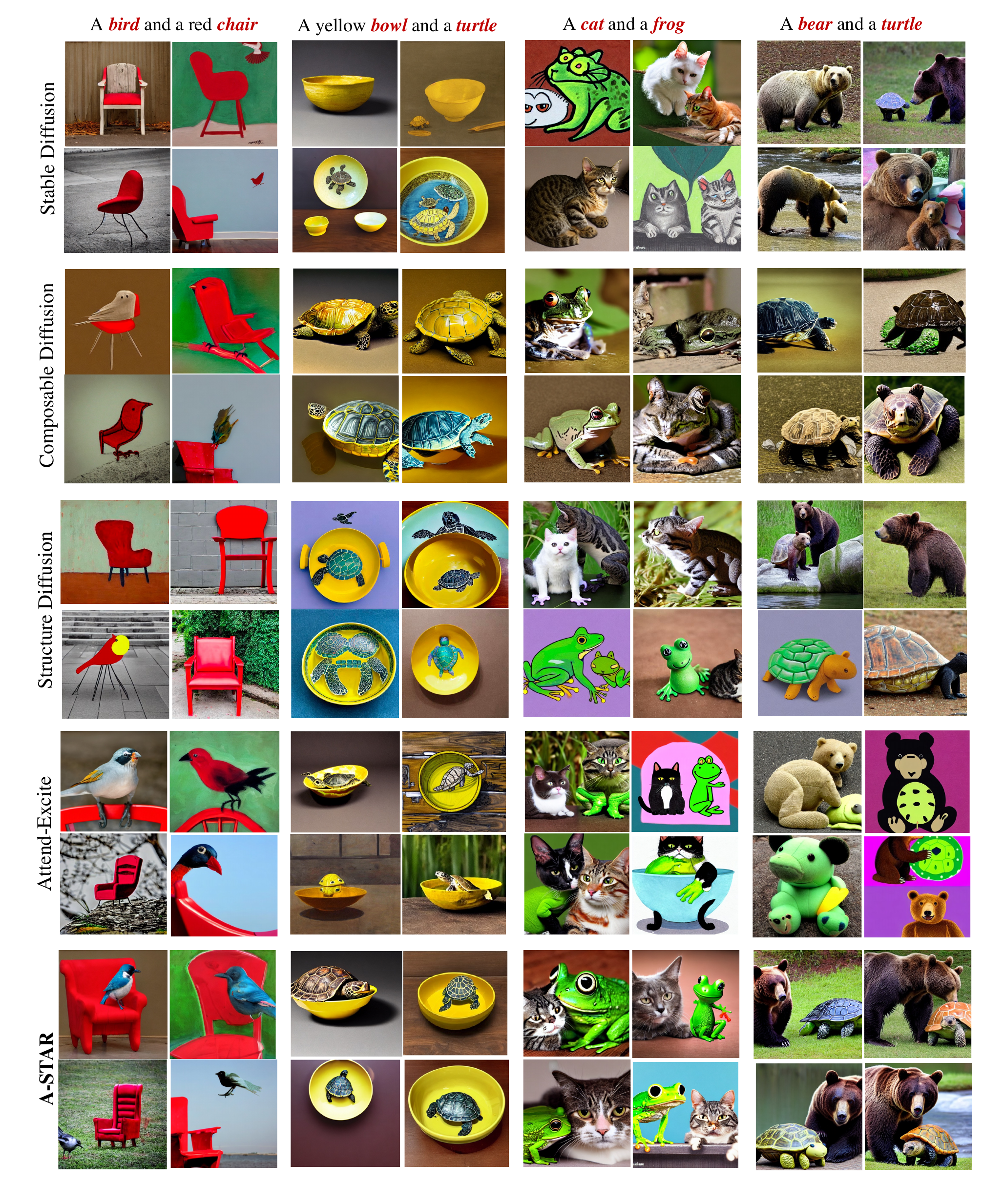}
    \vspace{-16pt}
    \caption{Comparison of A-STAR with recent state-of-the-art methods. For each prompt, we generate four images.}
    \vspace{-10pt}
    
    \label{fig:mainQualComparisonSOTA}
\end{figure*}

\begin{figure}
    \centering
    \includegraphics[width = \linewidth]{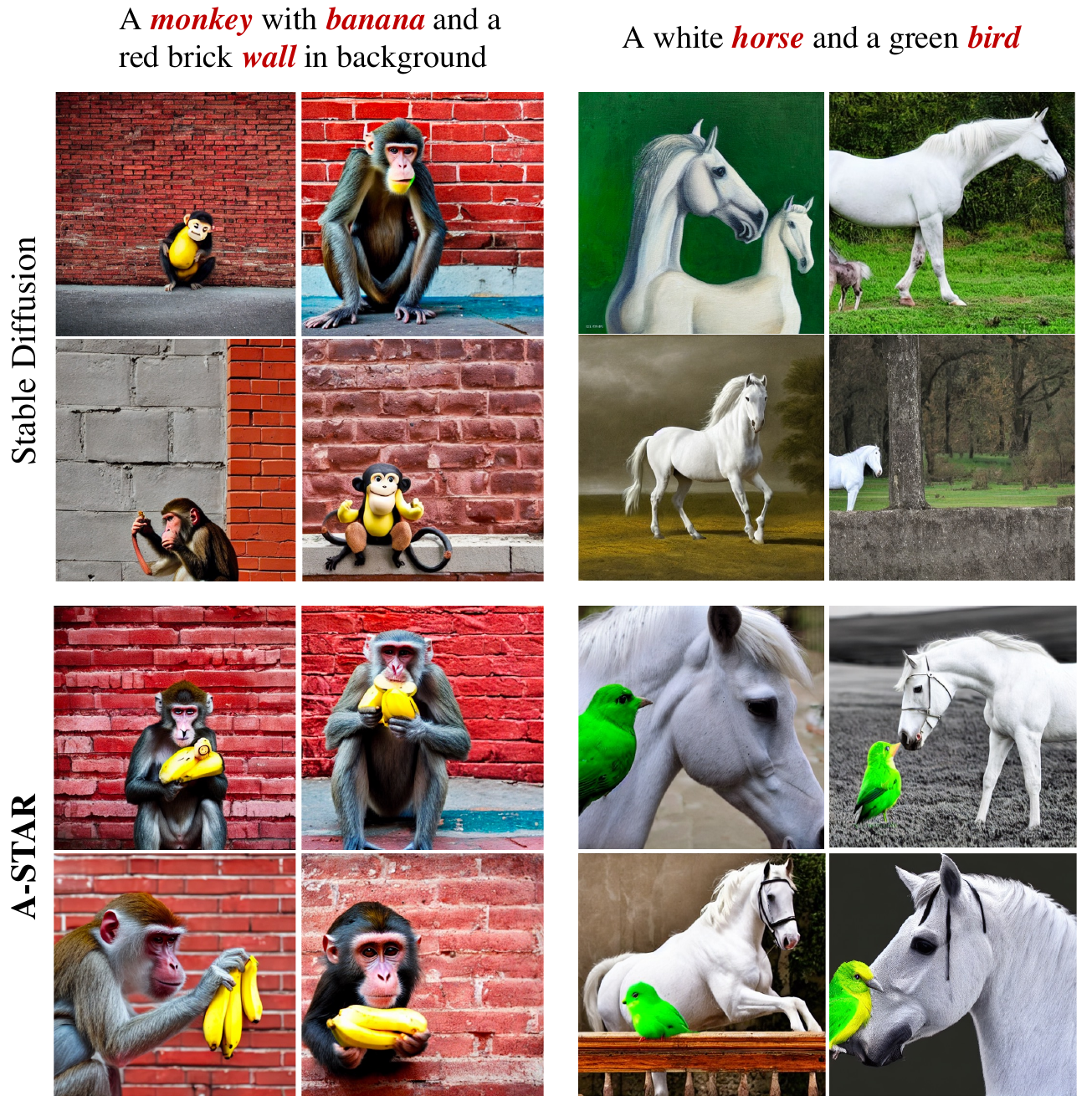}
    \vspace{-18pt}
    \caption{More comparisons with Stable Diffusion.}
    \vspace{-14pt}
    \label{fig:complex_prompts}
\end{figure}

\subsection{Attention Segregation Loss}

As discussed above, one key issue with existing work \cite{rombach2022high} is the overlap (in highly activated regions) between cross-attention maps of various concepts, resulting in confusion that leads to the model to skip several of them in the final generation. Recall the results in Figure~\ref{fig:teaser_qual} where baseline Stable Diffusion \cite{rombach2022high} misses out the \textit{ship} in the first example, the \textit{crown} in the second example, and the \textit{salmon} in the third example. Our key insight to address this issue is simple: eliminate this source of model confusion by reducing this attention overlap as much as possible. By doing so, we explicitly force the DDPM denoising process to have \textit{separate, highly activated} regions for each concept. This eventually leads to a  $\mathbf{z}_{0}$ that is representative of all concepts that can be decoded to the desirable image. We realize this idea with our attention segregation loss. This loss operates on a pair of cross-attention maps, one for each concept, at each time step $t$. In cases where there are more than two concepts, we aggregate this loss over all possible pairs from the set of concepts $\mathcal{C}$. For instance, in the second example of Figure~\ref{fig:teaser_qual}, we consider the \textit{frog}-\textit{crown} pair for this purpose. Given $\mathbf{A}_t^m$ and $\mathbf{A}_t^n$ to be a pair of cross-attention maps for concepts $m, n \in \mathcal{C}$ at the time step $t$, our proposed attention segregation loss is defined as: 

\begin{equation}
    \mathcal{L}_\text{seg} = \sum_{\substack{m, n \in \mathcal{C} \\ \forall m>n }} \left[ \dfrac{\sum_{ij}\text{min}([\mathbf{A}_t^m]_{ij}, [\mathbf{A}_t^n]_{ij})}{\sum_{ij}([\mathbf{A}_t^m]_{ij} + [\mathbf{A}_t^n]_{ij})} \right]
    \label{eq:attn_segregation}
\end{equation}

where $[\mathbf{A}_t^m]_{ij}$ is the pixel value at the $(i,j)$ location. At its core, the attention segregation loss seeks to segregate/separate the high-response regions for $\mathbf{A}_t^m$ and $\mathbf{A}_t^n$ by calculating and reducing their intersection-over-union (IoU) value. By aggregating over all pairs, this ensures all concepts have such separated regions. Some results with and without this loss are in Figure \ref{eq:attn_segregation}, where one can note improved separation between the \textit{bear} and \textit{turtle} attention maps which then leads to a more desirable output image.

\subsection{Attention Retention Loss}

The next contribution of our paper addresses the issue of attention decay discussed in Sections~\ref{sec:intro} and~\ref{sec:asr}. As discussed previously in Figure~\ref{fig:attention_retention}, the base Stable Diffusion model has information about all key concepts in the beginning of the denoising process but is not able to retain it by the time we reach $t=0$. Our key insight to address this issue is to explicitly force the model to retain the information throughout the denoising process by means of consistency constraints. By doing so, we explicitly ensure the DDPM denoising process will produce a $\mathbf{z}_0$ that has information about all concepts and can be decoded to the desired image.

We realize the idea above with our proposed attention retention loss. Given the attention map $\mathbf{A}_t^m$ of concept $m\in \mathcal{C}$ at timestep $t$, we determine the pixel regions with high activations and binarize the result to obtain its binary mask $\mathbf{B}_{t}^m$. Given we seek the retention of information from time step $t$ to $t-1$, we use $\mathbf{B}_{t}^m$ as a proxy for ground truth and ensure high response regions in the next time step's attention map $\mathbf{A}_{t-1}^m$ are consistent with $\mathbf{B}_{t}^m$. This can be formalized as a simple IoU maximization objective between $\mathbf{A}_{t-1}^m$ and $\mathbf{B}_{t}^m$. We repeat this for all the concepts and aggregate the resulting loss, giving us our proposed attention retention loss:

\vspace{-12pt}
\begin{equation}
    \mathcal{L}_\text{ret} = \sum_{m \in \mathcal{C}} \left[ 1 -  \dfrac{\sum_{ij}\text{min}([\mathbf{A}_{t-1}^m]_{ij}, [\mathbf{B}_{t}^m]_{ij})}{\sum_{ij}([\mathbf{A}_{t-1}^m]_{ij} + [\mathbf{B}_{t}^m]_{ij})} \right]
    \label{eq:attn_retention}
\end{equation}

where as before $[\mathbf{A}_{t-1}^m]_{ij}$ is the pixel value at the $(i,j)$ location. By seeking to maximize the IoU between $\mathbf{A}_{t-1}^m$ and $\mathbf{B}_{t}^m$, we force the DDPM process to retain information from previous time steps, thereby alleviating the attention decay issue. Note that the binary mask $\mathbf{B}_{t}^m$ in Equation~\ref{eq:attn_retention} is updated at each time step. We show some results before and after this loss in the A-STAR row in Figure~\ref{fig:attention_retention} where one can note substantially improved information retention (e.g., highly activated \textit{umbrella} with our method vs. Stable Diffusion). Since we retain information for all concepts, the DDPM gives a $\mathbf{z}_0$ that captures all of \textit{dog}, \textit{beach}, and \textit{umbrella} in the final output when compared to the baseline.

\begin{figure*}
    \centering
    \includegraphics[width = \linewidth]{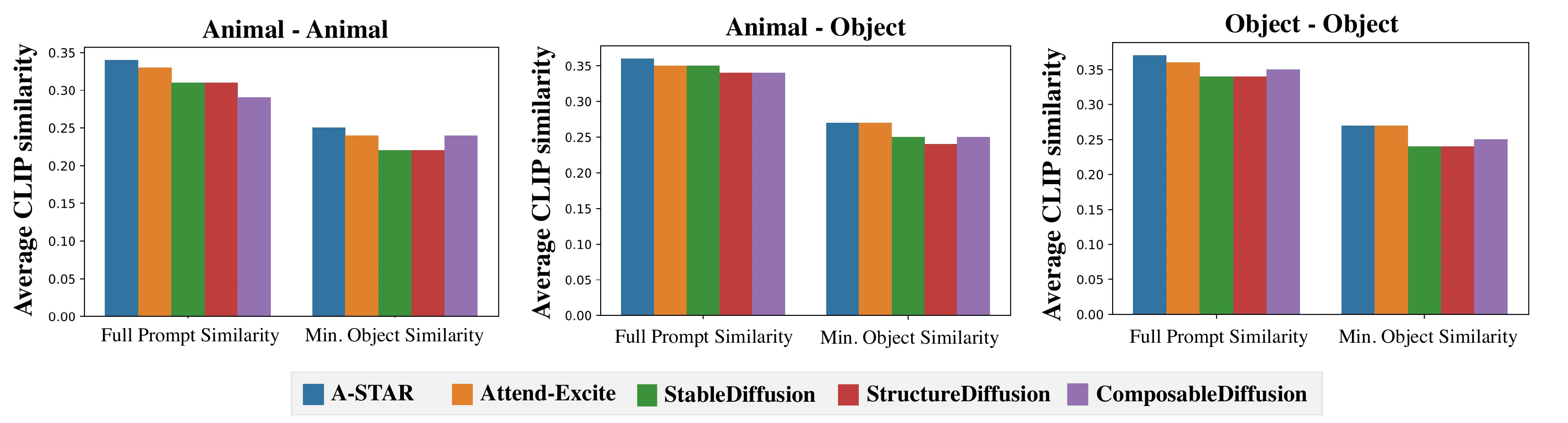}
    \vspace{-20pt}
    \caption{Average CLIP image-text similarities between the text prompts and the images generated by each method}
    \vspace{-10pt}
    \label{fig:graph_clip_it}
\end{figure*}

\subsection{Optimizing LDMs with A-STAR}
Our overall loss function (see also Figure~\ref{fig:placeholder}) includes both attention separation and attention retention loss as:
\vspace{-6pt}
\begin{equation}
   \mathcal{L}=\mathcal{L}_\text{seg} + \mathcal{L}_\text{ret}
  \label{eq:total_loss}
\end{equation}
\vspace{-16pt}

We now only need to direct the latent code at the current time step $\mathbf{z}_{t}$ in the right direction as measured by this overall loss (see the intuition for our proposed losses again in Section~\ref{sec:asr}). We realize this with a latent update: $\mathbf{z}_t' = \mathbf{z}_t - \alpha_t \cdot \nabla_{\mathbf{z}_t} \mathcal{L}$, where $\alpha_t$ is the step size of the gradient update. This updated $\mathbf{z}_t'$ is then used in the next denoising step to obtain $\mathbf{z}_{t-1}$, which is then again updated in a similar fashion (and the process repeats until the last step).

\section{Results}
\label{sec:results}
Given the unavailability of standard benchmarks to evaluate text-to-image models, we use a mix of commonly used prompts for qualitative evaluation and the protocol in prior work \cite{chefer2023attend} for a quantitative evaluation. In particular, this involves constructing prompts with two subjects in the following fashion: $[$\textit{animalA}-\textit{animalB}$]$, $[$\textit{animal}-\textit{color}$]$, and $[$\textit{colorA; objectA}-\textit{colorB; objectB}$]$. We do evaluate on much more complex prompts as well (see Figures~\ref{fig:teaser_qual} and~\ref{fig:complex_prompts}).

\textbf{Qualitative Results.}
We first begin by discussing our generation outputs. In Figure~\ref{fig:mainQualComparisonSOTA}, we compare A-STAR's results with other recent competing methods like Composable Diffusion \cite{liu2022compositional}, Structure Diffusion \cite{feng2022training} and Attend-Excite \cite{chefer2023attend}, and one can note A-STAR clearly outperforms these techniques. For instance, in the first column, all of Stable Diffusion \cite{rombach2022high}, Composable Diffusion \cite{liu2022compositional}, and Structure Diffusion \cite{feng2022training} are not able to capture the two concepts of \textit{bird} and \textit{red chair} whereas A-STAR is able to, showing the importance of having two well segregated and activated attention regions that are retained across denoising steps. Further, as can be seen from the Attend-Excite \cite{chefer2023attend} results, the chair is either not fully visible (first two images) or is conflated into the bird (bottom-right), suggesting that simply maximizing patches in attention maps will not ensure the holistic properties of all concepts are captured. Similar observations can be made from results in the other three columns as well. We also observe significant qualitative improvements with A-STAR in the binding of attributes to these concepts. For instance, while competing methods do incorrect binding (e.g., many baseline generations have red attribute transferred to bird as well in column 1, the yellow attribute transferred to turtle as well in column 2), our method is able to resolve this correctly (e.g., only red chairs and yellow bowls in our results). 

\begin{wrapfigure}{r}{0.25\textwidth}
\includegraphics[width=0.25\textwidth]{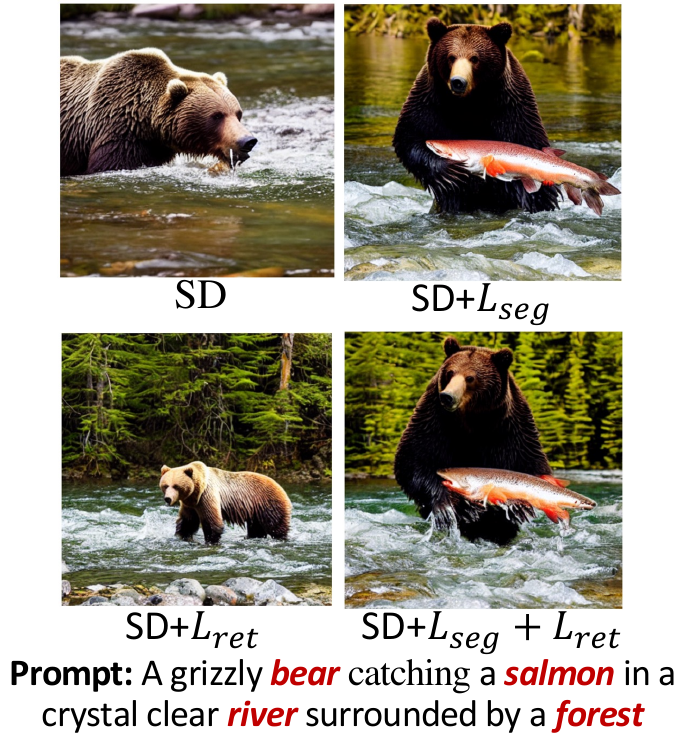} 
\caption{Ablation results.}
\label{fig:qual_results_ablation}
\end{wrapfigure} 

Next, in Figure~\ref{fig:complex_prompts}, we compare our results with baseline Stable Diffusion \cite{rombach2022high} on more complex prompts. In the first column, while the baseline is either missing one of the three concepts (e.g., \textit{banana} in many cases) or mixing them up (e.g., \textit{monkey} in \textit{banana} appearance), A-STAR captures all of them as desired. Similarly, A-STAR generates both the \textit{white horse} and the \textit{green bird} as desired. We end this section with some qualitative ablation results that demonstrate the impact of our losses. In Figure~\ref{fig:qual_results_ablation}, we show the result of incrementally adding our losses to base Stable Diffusion \cite{rombach2022high}. When we add $\mathcal{L}_{seg}$, the \textit{salmon} shows up since it was previously omitted due to cross-attention overlap between \textit{salmon} and \textit{bear}. When we add $\mathcal{L}_{ret}$, the \textit{forest} shows up since its information was present in the first few timesteps and hence retained with $\mathcal{L}_{ret}$. Finally, we achieve both these aspects as expected with $\mathcal{L}_{seg}$ and $\mathcal{L}_{ret}$.

\begin{table}[]
\resizebox{0.48\textwidth}{!}{%
\begin{tabular}{@{}l|ccc@{}}
\toprule
Method                                 & Animal - Animal                 & Animal - Object                & Object - Object                \\ \midrule
Stable \cite{rombach2022high}          & 0.76 \textcolor{red}{(-7.9\%)}  & 0.78 \textcolor{red}{(-7.7\%)} & 0.77 \textcolor{red}{(-6.5\%)} \\
Composable \cite{liu2022compositional} & 0.69 \textcolor{red}{(-18.9\%)} & 0.77 \textcolor{red}{(-9.1\%)} & 0.76 \textcolor{red}{(-7.9\%)} \\
Structure \cite{feng2022training}      & 0.76 \textcolor{red}{(-7.9\%)}  & 0.78 \textcolor{red}{(-7.7\%)} & 0.76 \textcolor{red}{(-7.9\%)} \\
Attend-Excite \cite{chefer2023attend}  & 0.80 \textcolor{red}{(-2.5\%)}  & 0.82 \textcolor{red}{(-2.4\%)} & 0.81 \textcolor{red}{(-1.2\%)} \\ \midrule
\textbf{A-STAR}                      & \textbf{0.82}                   & \textbf{0.84}                  & \textbf{0.82}                  \\ \bottomrule
\end{tabular}%
}

\vspace{-8pt}
    \caption{Text-text similarities between the text prompts and BLIP-generated captions over the generated images.}
    \vspace{-4pt}
    \label{tab:qual_text_text}

\end{table}

\begin{table}[]
\resizebox{0.48\textwidth}{!}{%
\begin{tabular}{@{}l|ccc@{}}
\toprule
Method                                              & Animal - Animal                          & Animal - Object                          & Object - Object                          \\ \midrule
Stable \cite{rombach2022high}                       & 0.76                                     & 0.78                                     & 0.77                                     \\
Stable + $\mathcal{L}_{ret}$                        & 0.78 \textcolor{blue}{(+2.5\%)}          & 0.83 \textcolor{blue}{(+6.4\%)}          & 0.79 \textcolor{blue}{(+2.6\%)}          \\
Stable + $\mathcal{L}_{seg}$                        & 0.79 \textcolor{blue}{(+4.0\%)}          & 0.82 \textcolor{blue}{(+5.1\%)}          & 0.80 \textcolor{blue}{(+3.9\%)}          \\
Stable + $\mathcal{L}_{ret}$ +  $\mathcal{L}_{seg}$ & \textbf{0.82} \textcolor{blue}{(+7.9\%)} & \textbf{0.84} \textcolor{blue}{(+7.7\%)} & \textbf{0.82} \textcolor{blue}{(+6.5\%)} \\ \bottomrule
\end{tabular}%
}
    \vspace{-8pt}
    \caption{Ablation results for text-text similarities.}
    \vspace{-10pt}
    \label{tab:abl_text_text}
\end{table}

\textbf{Quantitative Results.} We follow existing protocol \cite{chefer2023attend} and quantify performance with CLIP \cite{radford2021learning} distances. We first generate 64 images with randomly selected seeds and compute the  average image-text  cosine similarity using CLIP for each prompt. Here, as in prior work \cite{chefer2023attend}, we use both full prompt similarity (i.e., cosine similarity between full prompt and generated image) and minimum object similarity (i.e., minimum of the two similarities between generated image and each of the two subject prompts) and report results in Figure~\ref{fig:graph_clip_it}. A-STAR outperforms the baselines across both metrics across all the three categories. In particular, it outperforms Attend-Excite \cite{chefer2023attend} by $2.9\%$ and $1.4\%$ across all three subsets for full prompt similarity and minimum object similarity respectively (the corresponding improvements over Stable Diffusion \cite{rombach2022high} are $7.1\%$ and $10.8\%$).

We also compute text-text similarities by captioning the generated images with BLIP \cite{li2022blip} and comparing them with the input prompt. See Table~\ref{tab:qual_text_text} where much higher similarities with A-STAR is indicative of the semantic correctness of the generated results with our method. In the supplementary material document, we provide additional text-text similarity results with other standard metrics as well.
Finally, we also quantify the impact of each of our losses in Table~\ref{tab:abl_text_text} where we compute text-text similarities as above. While each loss individually improves baseline  \cite{rombach2022high} performance, the full model achieves the highest improvement, indicating the losses' complementarity. A graph similar to Figure~\ref{fig:graph_clip_it} for this ablation experiment can be found in the supplementary material.

\textbf{User study.} Finally, we conduct a user study with the generated images where we ask survey respondents to select which set of images (among sets from three different methods, see Table~\ref{tab:userStudy}) best represents the input text semantically. We randomly sample five prompts for each of the \textit{animal-animal}, \textit{animal-object}, and \textit{object-object} categories in Table~\ref{tab:userStudy} and generate four different images for each prompt using each of Stable Diffusion \cite{rombach2022high}, Attend-Excite \cite{chefer2023attend}, and A-STAR. From Table~\ref{tab:userStudy}, our method's results are preferred by a majority of the survey respondents, thus providing additional evidence for the impact of our proposed losses on the semantic faithfulness of the images generated by A-STAR. Supplementary material has more details.

\begin{table}[]
\resizebox{0.48\textwidth}{!}{%
\begin{tabular}{@{}c|ccc@{}}
\toprule
Method                                & Animal - Animal & Animal - Object & Object - Object \\ \midrule
Stable \cite{rombach2022high}         & $2.2\%$           & $6.7\%$           & $3.0\%$          \\
Attend-Excite \cite{chefer2023attend} & $3.0\%$           & $14.1\%$          & $13.3\%$          \\ \midrule
\textbf{A-STAR}                     & $\textbf{94.8\%}$ & $\textbf{79.2\%}$ & $\textbf{83.7\%}$ \\ \bottomrule
\end{tabular}%
}
    \vspace{-8pt}
    \caption{Results from a user survey with $26$ respondents.}
    \vspace{-4pt}
    \label{tab:userStudy}
\end{table}

\section{Summary}

In this work, we notice that several baseline diffusion models are not faithful in capturing all the concepts from an input prompt in the generated image, and identify two key issues that contributes to this behavior. First, in many cases, there is significant pixel overlap among concepts in their intermediate attention maps that leads to model confusion, and second, these models are not able to retain knowledge (in attention maps) across all denoising timesteps. We propose two loss functions, attention segregation and attention retention, that fixes these issues directly at inference time, without any retraining. We conduct extensive qualitative experiments with a variety of prompts and demonstrate that the images are substantially more semantically faithful to the input prompts, when compared to many recently proposed models. Further, we also quantify our improvements with protocols from the literature as well as a user survey, which clearly brings out the efficacy of A-STAR.

{\small
\bibliographystyle{ieee_fullname}
\bibliography{egbib}

\begin{thebibliography}{10}\itemsep=-1pt

\bibitem{brooks2022instructpix2pix}
Tim Brooks, Aleksander Holynski, and Alexei~A Efros.
\newblock Instructpix2pix: Learning to follow image editing instructions.
\newblock {\em arXiv preprint arXiv:2211.09800}, 2022.

\bibitem{chefer2023attend}
Hila Chefer, Yuval Alaluf, Yael Vinker, Lior Wolf, and Daniel Cohen-Or.
\newblock Attend-and-excite: Attention-based semantic guidance for
  text-to-image diffusion models.
\newblock {\em arXiv preprint arXiv:2301.13826}, 2023.

\bibitem{feng2022training}
Weixi Feng, Xuehai He, Tsu-Jui Fu, Varun Jampani, Arjun Akula, Pradyumna
  Narayana, Sugato Basu, Xin~Eric Wang, and William~Yang Wang.
\newblock Training-free structured diffusion guidance for compositional
  text-to-image synthesis.
\newblock {\em arXiv preprint arXiv:2212.05032}, 2022.

\bibitem{hao2022optimizing}
Yaru Hao, Zewen Chi, Li Dong, and Furu Wei.
\newblock Optimizing prompts for text-to-image generation.
\newblock {\em arXiv preprint arXiv:2212.09611}, 2022.

\bibitem{hertz2022prompt}
Amir Hertz, Ron Mokady, Jay Tenenbaum, Kfir Aberman, Yael Pritch, and Daniel
  Cohen-Or.
\newblock Prompt-to-prompt image editing with cross attention control.
\newblock {\em arXiv preprint arXiv:2208.01626}, 2022.

\bibitem{ho2022classifier}
Jonathan Ho and Tim Salimans.
\newblock Classifier-free diffusion guidance.
\newblock {\em arXiv preprint arXiv:2207.12598}, 2022.

\bibitem{huang2018introvae}
Huaibo Huang, Ran He, Zhenan Sun, Tieniu Tan, et~al.
\newblock Introvae: Introspective variational autoencoders for photographic
  image synthesis.
\newblock {\em Advances in neural information processing systems}, 31, 2018.

\bibitem{isola2017image}
Phillip Isola, Jun-Yan Zhu, Tinghui Zhou, and Alexei~A Efros.
\newblock Image-to-image translation with conditional adversarial networks.
\newblock In {\em Proceedings of the IEEE conference on computer vision and
  pattern recognition}, pages 1125--1134, 2017.

\bibitem{karras2019style}
Tero Karras, Samuli Laine, and Timo Aila.
\newblock A style-based generator architecture for generative adversarial
  networks.
\newblock In {\em Proceedings of the IEEE/CVF conference on computer vision and
  pattern recognition}, pages 4401--4410, 2019.

\bibitem{kingma2013auto}
Diederik~P Kingma and Max Welling.
\newblock Auto-encoding variational bayes.
\newblock {\em arXiv preprint arXiv:1312.6114}, 2013.

\bibitem{li2022blip}
Junnan Li, Dongxu Li, Caiming Xiong, and Steven Hoi.
\newblock Blip: Bootstrapping language-image pre-training for unified
  vision-language understanding and generation.
\newblock In {\em International Conference on Machine Learning}, pages
  12888--12900. PMLR, 2022.

\bibitem{li2023gligen}
Yuheng Li, Haotian Liu, Qingyang Wu, Fangzhou Mu, Jianwei Yang, Jianfeng Gao,
  Chunyuan Li, and Yong~Jae Lee.
\newblock Gligen: Open-set grounded text-to-image generation.
\newblock {\em arXiv preprint arXiv:2301.07093}, 2023.

\bibitem{liu2022compositional}
Nan Liu, Shuang Li, Yilun Du, Antonio Torralba, and Joshua~B Tenenbaum.
\newblock Compositional visual generation with composable diffusion models.
\newblock In {\em Computer Vision--ECCV 2022: 17th European Conference, Tel
  Aviv, Israel, October 23--27, 2022, Proceedings, Part XVII}, pages 423--439.
  Springer, 2022.

\bibitem{liu2022design}
Vivian Liu and Lydia~B Chilton.
\newblock Design guidelines for prompt engineering text-to-image generative
  models.
\newblock In {\em Proceedings of the 2022 CHI Conference on Human Factors in
  Computing Systems}, pages 1--23, 2022.

\bibitem{mokady2022null}
Ron Mokady, Amir Hertz, Kfir Aberman, Yael Pritch, and Daniel Cohen-Or.
\newblock Null-text inversion for editing real images using guided diffusion
  models.
\newblock {\em arXiv preprint arXiv:2211.09794}, 2022.

\bibitem{nichol2021glide}
Alex Nichol, Prafulla Dhariwal, Aditya Ramesh, Pranav Shyam, Pamela Mishkin,
  Bob McGrew, Ilya Sutskever, and Mark Chen.
\newblock Glide: Towards photorealistic image generation and editing with
  text-guided diffusion models.
\newblock {\em arXiv preprint arXiv:2112.10741}, 2021.

\bibitem{park2019semantic}
Taesung Park, Ming-Yu Liu, Ting-Chun Wang, and Jun-Yan Zhu.
\newblock Semantic image synthesis with spatially-adaptive normalization.
\newblock In {\em Proceedings of the IEEE/CVF conference on computer vision and
  pattern recognition}, pages 2337--2346, 2019.

\bibitem{radford2021learning}
Alec Radford, Jong~Wook Kim, Chris Hallacy, Aditya Ramesh, Gabriel Goh,
  Sandhini Agarwal, Girish Sastry, Amanda Askell, Pamela Mishkin, Jack Clark,
  et~al.
\newblock Learning transferable visual models from natural language
  supervision.
\newblock In {\em International conference on machine learning}, pages
  8748--8763. PMLR, 2021.

\bibitem{raffel2020exploring}
Colin Raffel, Noam Shazeer, Adam Roberts, Katherine Lee, Sharan Narang, Michael
  Matena, Yanqi Zhou, Wei Li, and Peter~J Liu.
\newblock Exploring the limits of transfer learning with a unified text-to-text
  transformer.
\newblock {\em The Journal of Machine Learning Research}, 21(1):5485--5551,
  2020.

\bibitem{ramesh2022hierarchical}
Aditya Ramesh, Prafulla Dhariwal, Alex Nichol, Casey Chu, and Mark Chen.
\newblock Hierarchical text-conditional image generation with clip latents.
\newblock {\em arXiv preprint arXiv:2204.06125}, 2022.

\bibitem{reimers2019sentence}
Nils Reimers and Iryna Gurevych.
\newblock Sentence-bert: Sentence embeddings using siamese bert-networks.
\newblock {\em arXiv preprint arXiv:1908.10084}, 2019.

\bibitem{rombach2022high}
Robin Rombach, Andreas Blattmann, Dominik Lorenz, Patrick Esser, and Bj{\"o}rn
  Ommer.
\newblock High-resolution image synthesis with latent diffusion models.
\newblock In {\em Proceedings of the IEEE/CVF Conference on Computer Vision and
  Pattern Recognition}, pages 10684--10695, 2022.

\bibitem{ronneberger2015u}
Olaf Ronneberger, Philipp Fischer, and Thomas Brox.
\newblock U-net: Convolutional networks for biomedical image segmentation.
\newblock In {\em Medical Image Computing and Computer-Assisted
  Intervention--MICCAI 2015: 18th International Conference, Munich, Germany,
  October 5-9, 2015, Proceedings, Part III 18}, pages 234--241. Springer, 2015.

\bibitem{saharia2022photorealistic}
Chitwan Saharia, William Chan, Saurabh Saxena, Lala Li, Jay Whang, Emily
  Denton, Seyed Kamyar~Seyed Ghasemipour, Burcu~Karagol Ayan, S~Sara Mahdavi,
  Rapha~Gontijo Lopes, et~al.
\newblock Photorealistic text-to-image diffusion models with deep language
  understanding.
\newblock {\em arXiv preprint arXiv:2205.11487}, 2022.

\bibitem{tao2022df}
Ming Tao, Hao Tang, Fei Wu, Xiao-Yuan Jing, Bing-Kun Bao, and Changsheng Xu.
\newblock Df-gan: A simple and effective baseline for text-to-image synthesis.
\newblock In {\em Proceedings of the IEEE/CVF Conference on Computer Vision and
  Pattern Recognition}, pages 16515--16525, 2022.

\bibitem{van2017neural}
Aaron Van Den~Oord, Oriol Vinyals, et~al.
\newblock Neural discrete representation learning.
\newblock {\em Advances in neural information processing systems}, 30, 2017.

\bibitem{vaswani2017attention}
Ashish Vaswani, Noam Shazeer, Niki Parmar, Jakob Uszkoreit, Llion Jones,
  Aidan~N Gomez, {\L}ukasz Kaiser, and Illia Polosukhin.
\newblock Attention is all you need.
\newblock {\em Advances in neural information processing systems}, 30, 2017.

\bibitem{wang2022diffusiondb}
Zijie~J Wang, Evan Montoya, David Munechika, Haoyang Yang, Benjamin Hoover, and
  Duen~Horng Chau.
\newblock Diffusiondb: A large-scale prompt gallery dataset for text-to-image
  generative models.
\newblock {\em arXiv preprint arXiv:2210.14896}, 2022.

\bibitem{witteveen2022investigating}
Sam Witteveen and Martin Andrews.
\newblock Investigating prompt engineering in diffusion models.
\newblock {\em arXiv preprint arXiv:2211.15462}, 2022.

\bibitem{xu2018attngan}
Tao Xu, Pengchuan Zhang, Qiuyuan Huang, Han Zhang, Zhe Gan, Xiaolei Huang, and
  Xiaodong He.
\newblock Attngan: Fine-grained text to image generation with attentional
  generative adversarial networks.
\newblock In {\em Proceedings of the IEEE conference on computer vision and
  pattern recognition}, pages 1316--1324, 2018.

\bibitem{yang2022reco}
Zhengyuan Yang, Jianfeng Wang, Zhe Gan, Linjie Li, Kevin Lin, Chenfei Wu, Nan
  Duan, Zicheng Liu, Ce Liu, Michael Zeng, et~al.
\newblock Reco: Region-controlled text-to-image generation.
\newblock {\em arXiv preprint arXiv:2211.15518}, 2022.

\bibitem{ye2021improving}
Hui Ye, Xiulong Yang, Martin Takac, Rajshekhar Sunderraman, and Shihao Ji.
\newblock Improving text-to-image synthesis using contrastive learning.
\newblock {\em arXiv preprint arXiv:2107.02423}, 2021.

\bibitem{zeng2022scenecomposer}
Yu Zeng, Zhe Lin, Jianming Zhang, Qing Liu, John Collomosse, Jason Kuen, and
  Vishal~M Patel.
\newblock Scenecomposer: Any-level semantic image synthesis.
\newblock {\em arXiv preprint arXiv:2211.11742}, 2022.

\bibitem{zhang2021cross}
Han Zhang, Jing~Yu Koh, Jason Baldridge, Honglak Lee, and Yinfei Yang.
\newblock Cross-modal contrastive learning for text-to-image generation.
\newblock In {\em Proceedings of the IEEE/CVF conference on computer vision and
  pattern recognition}, pages 833--842, 2021.

\bibitem{zhang2023adding}
Lvmin Zhang and Maneesh Agrawala.
\newblock Adding conditional control to text-to-image diffusion models.
\newblock {\em arXiv preprint arXiv:2302.05543}, 2023.

\bibitem{zhu2017unpaired}
Jun-Yan Zhu, Taesung Park, Phillip Isola, and Alexei~A Efros.
\newblock Unpaired image-to-image translation using cycle-consistent
  adversarial networks.
\newblock In {\em Proceedings of the IEEE international conference on computer
  vision}, pages 2223--2232, 2017.

\bibitem{zhu2017toward}
Jun-Yan Zhu, Richard Zhang, Deepak Pathak, Trevor Darrell, Alexei~A Efros,
  Oliver Wang, and Eli Shechtman.
\newblock Toward multimodal image-to-image translation.
\newblock {\em Advances in neural information processing systems}, 30, 2017.

\bibitem{zhu2019dm}
Minfeng Zhu, Pingbo Pan, Wei Chen, and Yi Yang.
\newblock Dm-gan: Dynamic memory generative adversarial networks for
  text-to-image synthesis.
\newblock In {\em Proceedings of the IEEE/CVF conference on computer vision and
  pattern recognition}, pages 5802--5810, 2019.

\end{thebibliography}
}

\clearpage

\begin{appendices}

\section{}
In Section~\ref{sec:overlapDecay}, we show more results for both attention overlap and attention decay as well as provide evidence to show baseline Stable Diffusion \cite{rombach2022high} does not degrade with our losses in cases where it is already capturing all concepts in the prompt. In Section~\ref{sec:impl}, we give more implementation details. In Section~\ref{sec:addQuant}, we provide additional quantitative results where we report results of an ablation experiment that calculates CLIP image-text similarities like Figure 9 in the main paper. We also report SentenceTransformer based text-text similarity scores in this section. In Section~\ref{sec:addQual}, we provide additional qualitative results comparing our method with Attend-Excite \cite{chefer2023attend} on top of baseline Stable Diffusion. In Section~\ref{sec:layoutExample}, we show one possible application of our attention retention loss in generating layout-constrained image outputs. Finally, we conclude with some discussion on limitations of our method in Section~\ref{sec:limitations}.

\subsection{Attention Overlap and Attention Decay}
\label{sec:overlapDecay}
As discussed in the main paper, we identified two key issues with existing diffusion models: attention overlap and attention decay. Here, we show more examples. 

\begin{figure}
    \centering
    \includegraphics[width = \linewidth]{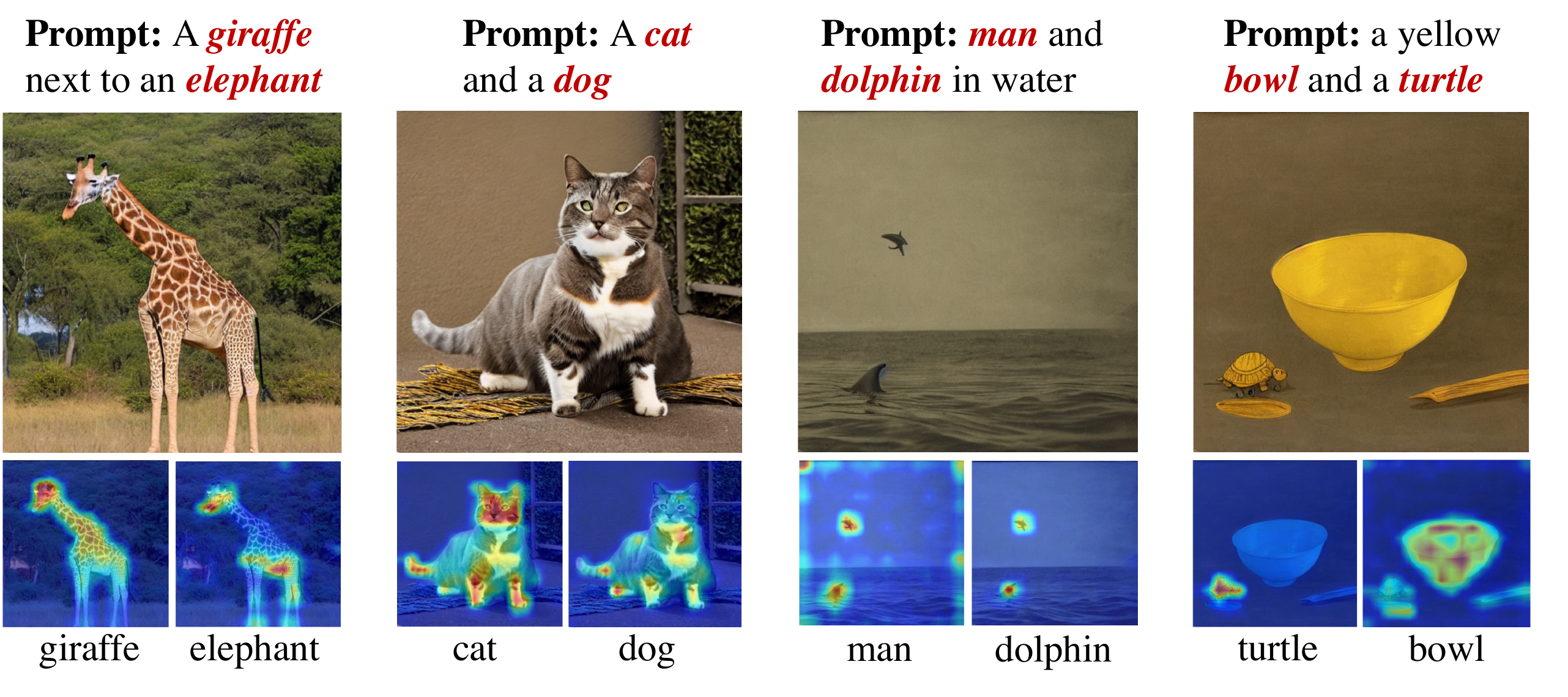}
    \caption{Examples demonstrating the issue of attention overlap in baseline Stable Diffusion}
    \label{fig:attention_overlap_base_sd}
\end{figure}

In Figure \ref{fig:attention_overlap_base_sd}, we demonstrate the issue of attention overlap with four examples. We notice that overlapping high-response regions in the attention maps lead to the \textit{elephant} getting missed in the generated output image in the first example, the \textit{dog} in the second example, and the \textit{man} in the third example. For instance, in the first example, there is significant overlap in the regions that correspond to high activations for both \textit{elephant} and \textit{giraffe} attention maps. Since they are highly activated in the same pixel regions, the final generated image is unable to
distinguish between the two subjects and is able to pick only one of the two. Similar reasoning follows for examples in columns 2 and 3. In column 4, we demonstrate the issue of incorrect attributes getting binded to the subjects due to attention overlap. Here, the attention map of \textit{bowl} has high responses for the same regions where the \textit{turtle} and other objects, leading to a mixup in the properties of the turtle and bowl (see final image where even the \textit{turtle} is yellow). 

\begin{figure*}
    \centering
    \includegraphics[width = \linewidth]{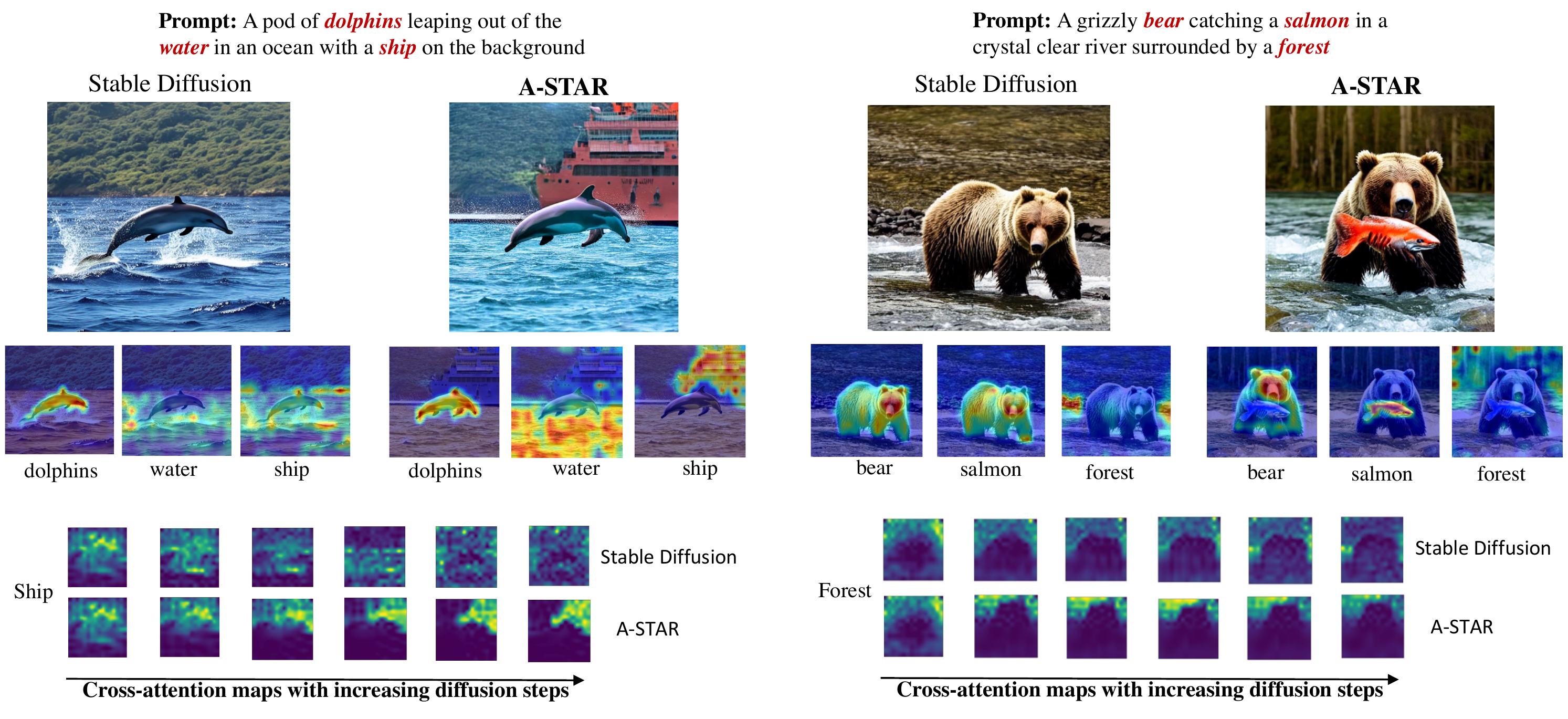}
    \caption{Examples demonstrating the issue of attention decay in baseline Stable Diffusion}
    \label{fig:attention_decay_base_sd}
\end{figure*}

In Figure \ref{fig:attention_decay_base_sd}, we show more examples to demonstrate the issue of attention decay. In the first example, one can note the \textit{ship} is missing in the baseline image output. Looking at the cross-attention maps for \textit{ship} across the denoising timesteps with baseline Stable Diffusion, it is clear that the information for this concept is present at the beginning but is not retained towards the end. Concretely, the pixel regions that were initially highly activated in the \textit{ship} attention map is very sparsely activated at the end. This results in the \textit{ship} not showing up in the final generation. A similar phenomenon can be observed with the \textit{forest} concept in the other example. Note that in both cases, with our proposed method (see A-STAR attention maps), we are able to correct this issue.

\begin{figure}
    \centering
    \includegraphics[width = 1.0\linewidth]{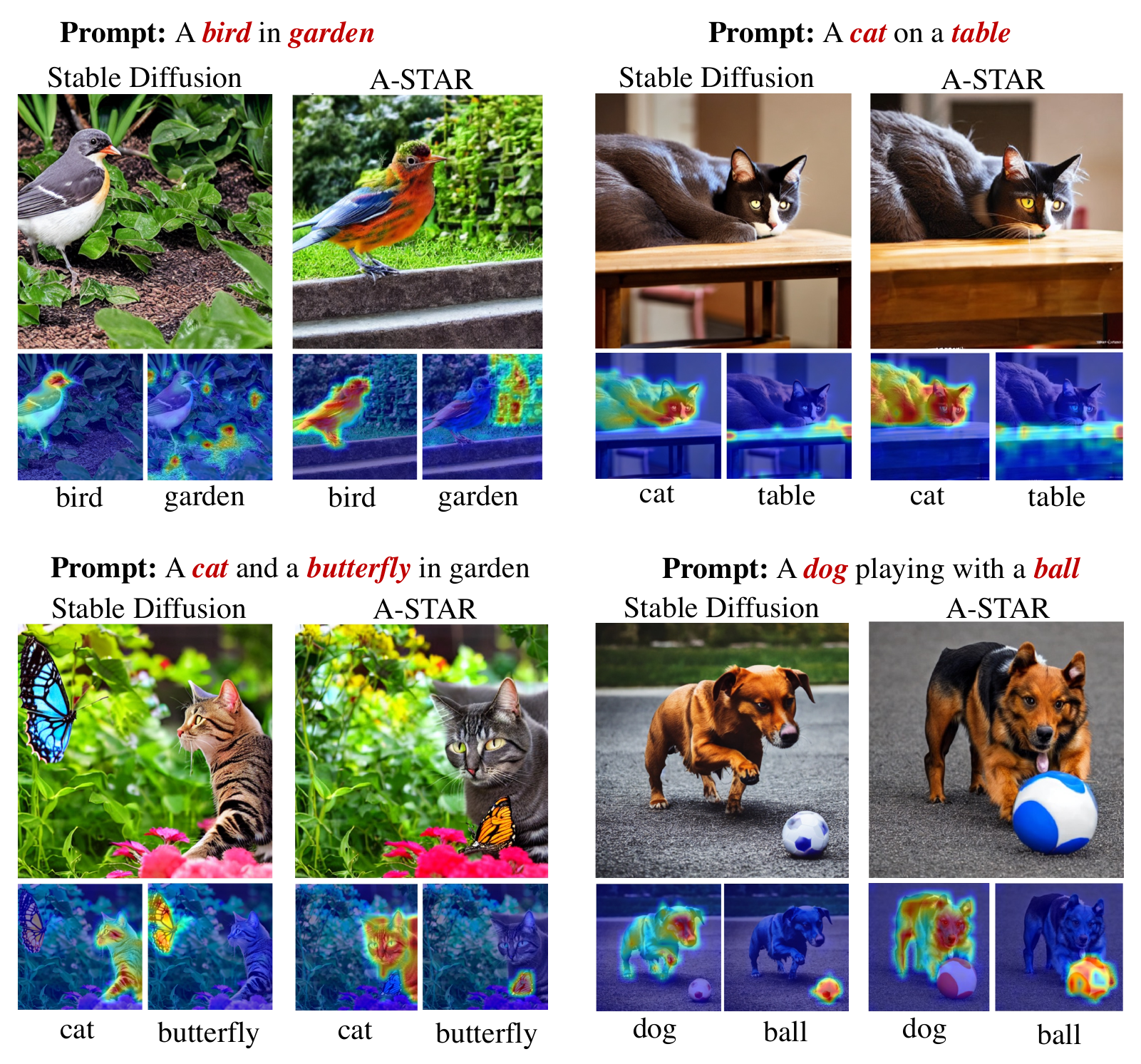}
    \caption{The first column shows the generated images and the corresponding attention maps for a set of prompts where baseline Stable Diffusion captures the text semantics well in the generation. In the second column, we show the generations for the same prompt and seed using A-STAR in order to demonstrate that A-STAR does not degrade the quality of generated images in terms of capturing semantics where baseline Stable Diffusion already works well.}
    \label{fig:attention_correct_base}
\end{figure}

In Figure \ref{fig:attention_correct_base}, we provide results for baseline Stable Diffusion and A-STAR for a set of prompts where the baseline model already captures the input semantics well. The motivation of this experiment is to show that in these cases, with our proposed losses, we are not degrading baseline performance. Let us consider the first example (top left) where the baseline model already has well-separated attention maps for \textit{bird} and \textit{garden}, resulting in both concepts being captured in the generated image. In this case, even after applying our losses with A-STAR, there is no degradation in the generated image and both concepts show up. Similarly, in the second example (top right), the baseline model has well-separated attention maps for \textit{cat} and \textit{table}, and this remains the case after applying the A-STAR losses, leading to both models giving the desired output. Similar observations can be made from the other two examples as well.

\subsection{Implementation Details}
\label{sec:impl}
Given an input text prompt, we consider all the possible subjects (e.g., nouns) while computing the two proposed losses. Let $\mathcal{C}$ denote the set of subjects identified given the prompt. To compute the losses $\mathcal{L}_{seg}$ and $\mathcal{L}_{ret}$, we first normalize the outputs of the cross-attention layers from the DDPM model to a range between $0$ and $1$ to obtain the attention maps $\mathbf{A}_t^m$ $\forall m \in C$. Note that $\mathcal{L}_{seg}$ considers all possible pairs of subjects present in the input text prompt. We next discuss how we compute the ground truth binary mask $\mathbf{B}_{t-1}$ used in $\mathcal{L}_{ret}$. Given the attention maps $\mathbf{A}_t^m$ for a subject $m$ at timestep $t$, we first determine a bounding box for the pixel regions with high activations and set all pixels within the bounding box to be 1 (and rest to 0), giving us the binary mask. Note that the mask computed at timestep $t$ gets utilised in the $\mathcal{L}_{ret}$ at timestep $t-1$.

\subsection{Additional Quantitative Results}
\label{sec:addQuant}
In the main paper in Fig 9, we showed CLIP image-text similarity comparisons with several existing diffusion models. In Fig 4 in this supplementary document, we show this graph for an ablation experiment to demonstrate the individual impact of our proposed losses. As can be seen from Fig 4 here, across all the three scenarios, while each of attention segregation and attention retention losses improve the performance of baseline Stable Diffusion, we obtain the best performance when both of them are used in conjunction.

In Table~\ref{tab:qual_text_text}, we show results corresponding to Table 1 in the main paper with cosine similarities computed using SentenceTransformer \cite{reimers2019sentence} embeddings. Specifically, we take the input prompt and the BLIP-generated caption, compute their respective  SentenceTransformerembeddings, and compute their cosine similarities. As can be seen from Table~\ref{tab:qual_text_text} here, A-STAR outperforms the baselines across all the three categories.

\begin{table}[]
\resizebox{0.48\textwidth}{!}{%
\begin{tabular}{@{}l|ccc@{}}
\toprule
Method                                 & Animal - Animal                 & Animal - Object                & Object - Object                \\ \midrule
Stable \cite{rombach2022high}          &  0.59  & 0.68  & 0.63  \\\
Attend-Excite \cite{chefer2023attend}  &  0.66  & 0.74  &  0.72 \\ \midrule
\textbf{A-STAR}                      & \textbf{0.68}                   & \textbf{0.75}                  & \textbf{0.73}                  \\ \bottomrule
\end{tabular}%
}

\vspace{-8pt}
    \caption{Text-text similarities between the text prompts and BLIP-generated captions over the generated images.}
    \vspace{-4pt}
    \label{tab:qual_text_text}

\end{table}

\subsection{Additional Qualitative Results}
\label{sec:addQual}

In Fig 5 here, we show more qualitative results comparing the performance of our proposed method with Attend-Excite on top of baseline Stable Diffusion. In each case, A-STAR gives more photorealistic imagery that captures all the input concepts. For instance, in the second column, A-STAR has both bear and turtle clearly captured in the final generation whereas both baseline Stable Diffusion and Attend-Excite fail. Similarly, in the fourth example, A-STAR generates both the horse and the bird whereas the other models either miss out one or both of these concepts. 

\subsection{Illustrating the Use of Attention Retention Loss in Layout-constrained Generation}
\label{sec:layoutExample}

\begin{figure}
    \centering
    \includegraphics[width = 1.0\linewidth]{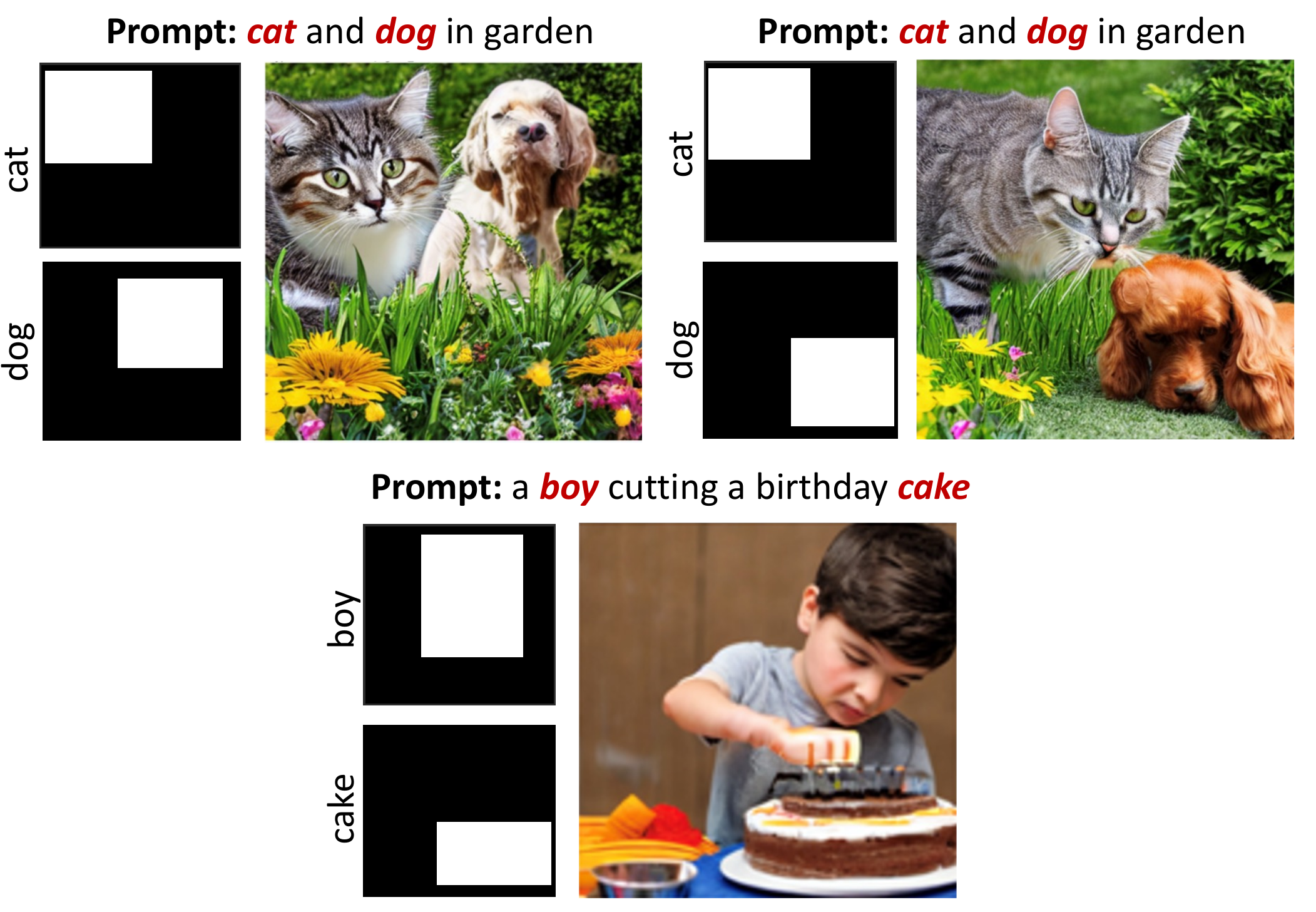}
    \caption{Example results to show an application of our attention retention loss where the ground-truth masks can come from a user in the form of layout constraints. In each of these cases, we can see the final concepts are generated in the spatial location represented by the corresponding input mask, e.g. cat at top-left and dog at bottom-right in the second example in the top right figure.}
    \label{fig:layout}
\end{figure}

An interesting application of our proposed attention retention loss is the ability to generate layout-constrained images. Instead of the automatically computed ground-truth masks as discussed in Section 1.2 above, these masks can be supplied directly by the user. More specifically, the users can delineate where they want certain concepts to show up in the final generation by means of binary layout maps. For instance, in the first example (top left) in Figure~\ref{fig:layout}, a user may specify the cat to show up at the top-left region in the image and the dog to show up in the middle-right region in the image. Given masks reflecting this layout constraint, we can apply our attention retention loss with these masks (instead of the ones in Section 1.2) and generate images constrained by these inputs. One can note in the results that this indeed is the case, with the cat being generated in top-left image region and the dog being generated in the middle-right region in the first example. Similar observations can be made from the other two examples.

\begin{figure*}
    \centering
    \includegraphics[width = \linewidth]{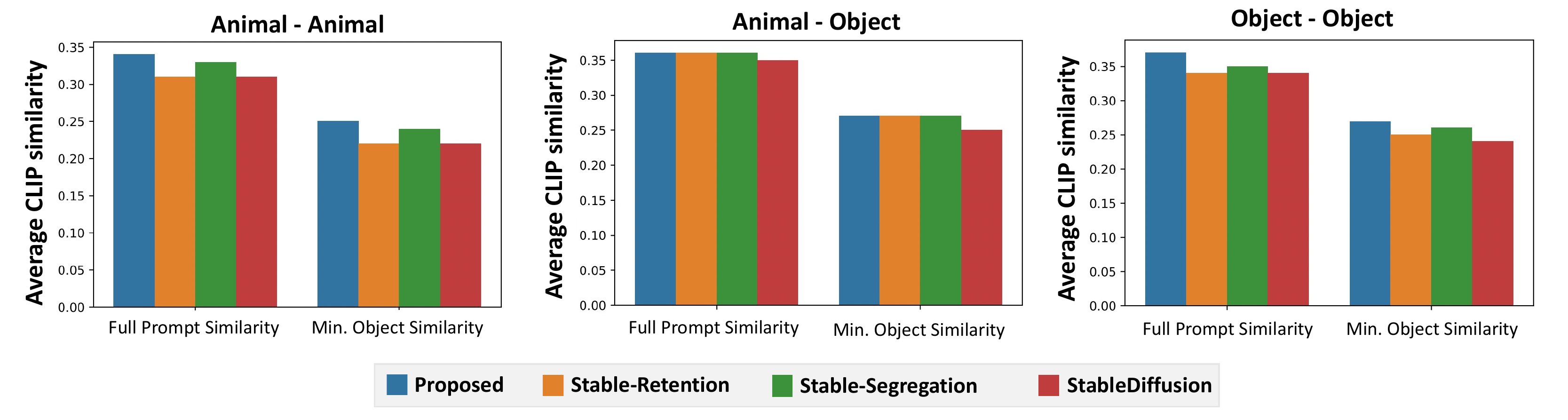}
    \caption{Ablation Study: Comparing Average CLIP image-text similarities between the text prompts and generated images}
    \label{fig:qual_results_all}
\end{figure*}

\begin{figure*}
    \centering
    \includegraphics[width = \linewidth]{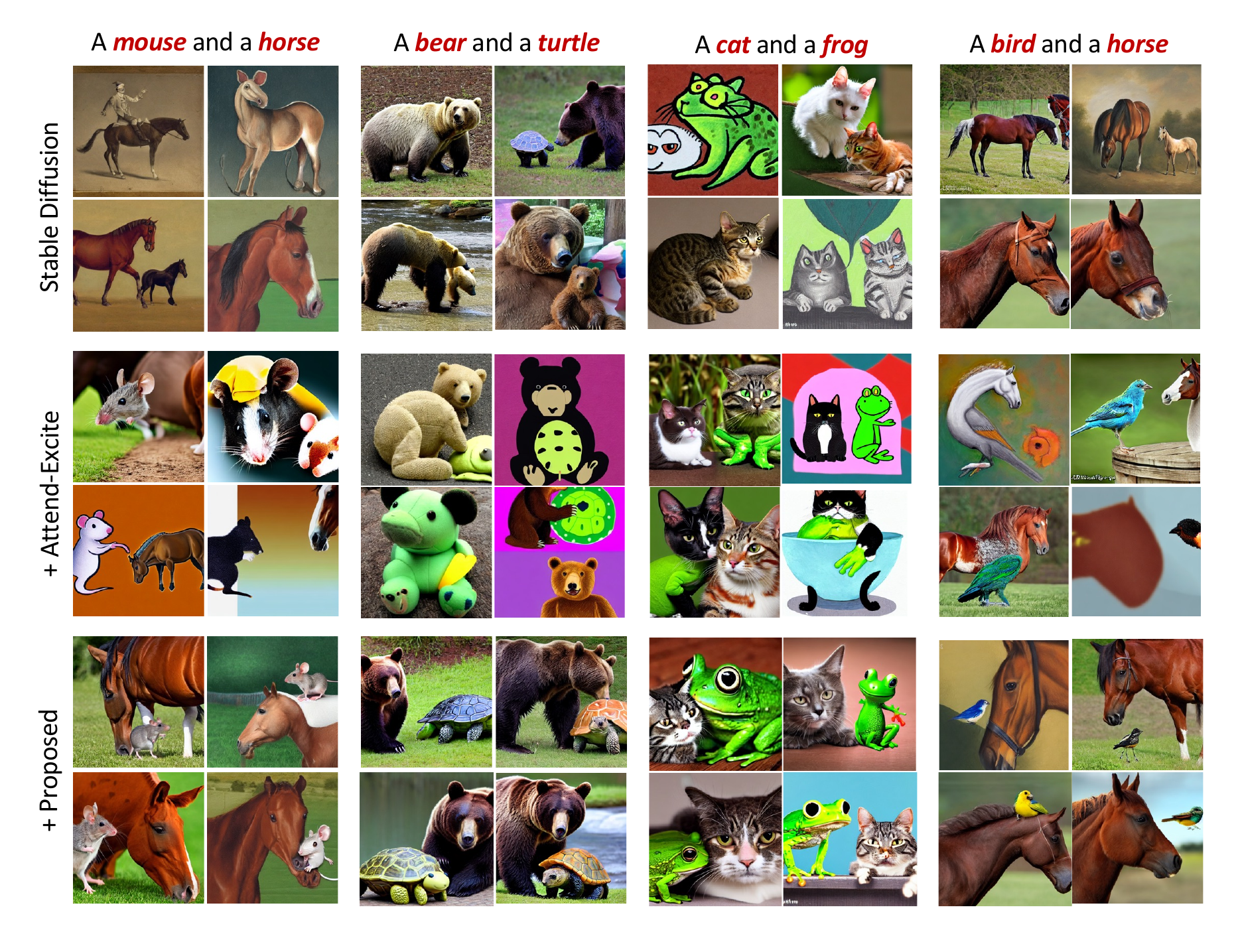}
    \caption{More comparison results of the proposed method vs Attend-Excite applied on top of base Stable Diffusion.}
    \label{fig:ours_vs_aae}
\end{figure*}

\subsection{Limitations}
\label{sec:limitations}

\begin{figure*}
    \centering
    \includegraphics[width = \linewidth]{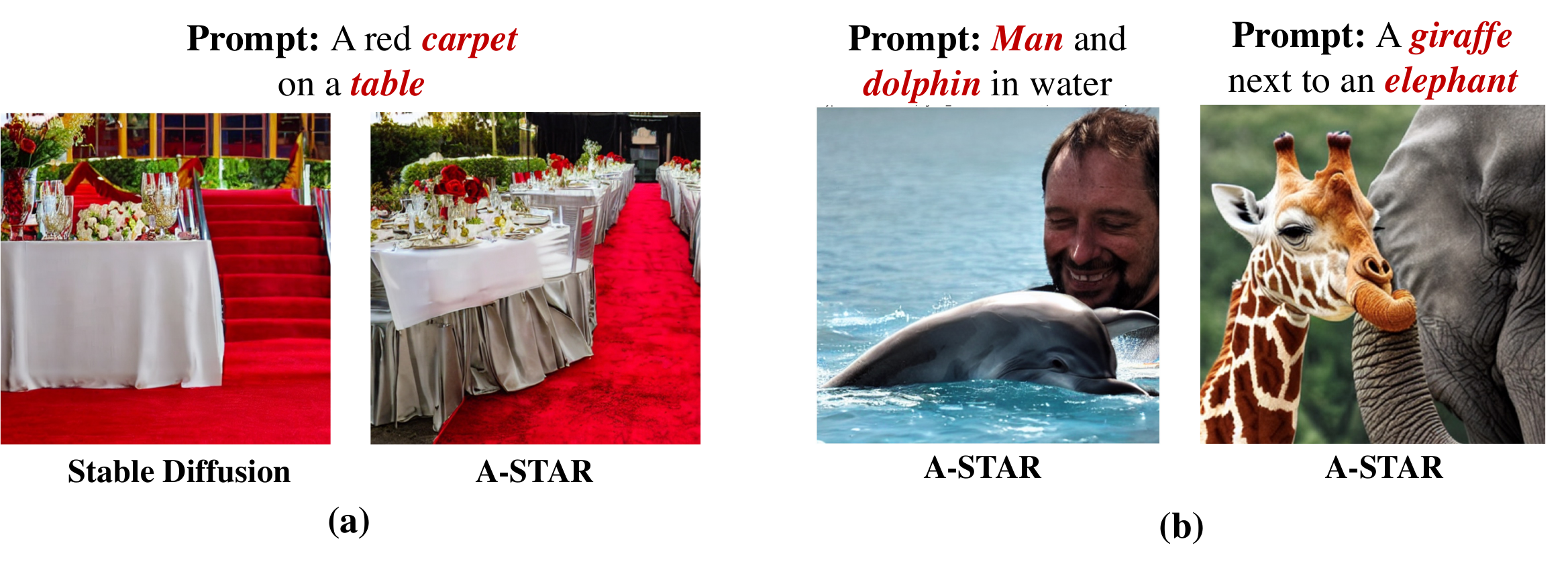}
    \caption{A-STAR limitations.}
    \label{fig:limitations}
\end{figure*}

In this section, we discuss a few limitations of our proposed method. In Figure~\ref{fig:limitations}(a), both the baseline model as well as A-STAR generate the \textit{red carpet} and the \textit{table} but lack an understanding of the relationship between the two concepts. In such cases, A-STAR is limited by the capabilities of the base model and as we discussed in both our proposed losses, we are currently not accounting for explicit relationship modeling between the concepts. However, given a computational model that captures these relationships, it can conceivably be added to our losses to reflect these relationships in the final output. 

In Figure~\ref{fig:limitations}(b), while A-STAR ensures both concepts (\textit{man} and \textit{dolphin} in first and \textit{giraffe} and \textit{elephant} in second) are captured in the final image, it may perhaps be more desirable to have these images generated at specific camera poses/viewpoints so as to capture these concepts more holistically. With advances in the ability to control diffusion model outputs \cite{zhang2023adding}, we can integrate our losses with such controlled generation techniques to improve these aspects.

\end{appendices}

\end{document}